\documentclass[10pt,twocolumn,letterpaper]{article}

\usepackage{cvpr}              %

\definecolor{cvprblue}{rgb}{0.21,0.49,0.74}
\usepackage[pagebackref,breaklinks,colorlinks,allcolors=cvprblue]{hyperref}
\usepackage{ulem}
\renewcommand{\emph}[1]{\textit{#1}}
\usepackage{colortbl,booktabs}
\usepackage{ragged2e}
\usepackage{makecell}
\usepackage{tabularx}
\usepackage{textcomp}
\usepackage{tabu} 
\usepackage{cuted}
\usepackage[shortlabels]{enumitem}
\usepackage{multirow}
\usepackage{setspace}

\usepackage{amsmath,amsfonts,bm}

\def\eqref#1{equation~\ref{#1}}

\def\1{\bm{1}}

\def\vk{{\bm{k}}}

\def\vo{{\bm{o}}}

\def\vq{{\bm{q}}}

\def\vu{{\bm{u}}}
\def\vv{{\bm{v}}}

\def\vx{{\bm{x}}}

\def\mW{{\bm{W}}}

\DeclareMathAlphabet{\mathsfit}{\encodingdefault}{\sfdefault}{m}{sl}
\SetMathAlphabet{\mathsfit}{bold}{\encodingdefault}{\sfdefault}{bx}{n}

\def\sR{{\mathbb{R}}}

\newcommand{\bx}{\mathbf{x}}

\newcommand{\bI}{\mathbf{I}}

\newcommand{\bepsilon}{{\boldsymbol{\epsilon}}}

\newcommand{\wan}{\texttt{Wan}\,}
\usepackage{arydshln}

\definecolor{c6}{HTML}{95Baa6}
\definecolor{c7}{HTML}{4b8dbc}

\definecolor{c5}{HTML}{C00000}

\definecolor{mycolor}{HTML}{a7caea} 
 
\definecolor{mygreen}{HTML}{95BAA6}
\definecolor{myblue}{HTML}{4b8dbc}
\definecolor{myyellow}{HTML}{E2CD89}
\newcommand{\yl}[1]{\textcolor{myyellow}{#1}}

\newcommand{\bl}[1]{\textcolor{myblue}{#1}}

\def\paperID{} %
\def\confName{CVPR}
\def\confYear{2026}

\title{\textsc{LinVideo}: A Post-Training Framework towards $\mathcal{O}(n)$ Attention in Efficient Video Generation}

\author{Yushi Huang\textsuperscript{1,3}\thanks{Work done during their internships at SenseTime Research.}\quad Xingtong Ge\textsuperscript{1} Ruihao Gong\textsuperscript{2,3}\thanks{Correspondence to: Ruihao Gong (\texttt{gongruihao@buaa.edu.cn}), Jun Zhang (\texttt{eejzhang@ust.hk}).}\quad Chengtao Lv\textsuperscript{3,4}\footnote[1]{}\quad Jun Zhang\textsuperscript{1}\footnote[2]{}\\
{\small \textsuperscript{1}Hong Kong University of Science and Technology \quad \textsuperscript{2}Beihang University \quad \textsuperscript{3}Sensetime Research \quad
\textsuperscript{4}Nanyang Technological University}
}

\begin{document}
\maketitle

\begin{abstract}

\noindent Video diffusion models (DMs) have enabled high-quality video synthesis, but their computation costs scale quadratically with sequence length due to the nature of self-attention. While linear attention offers a more efficient alternative, fully replacing quadratic attention demands costly pretraining. This is largely because linear attention lacks sufficient expressiveness and struggles with the complex spatiotemporal dynamics inherent to video generation. In this paper, we present \textsc{LinVideo}, an efficient data-free post-training framework that replaces a target number of self-attention modules with linear attention while preserving performance. First, we observe a significant disparity in the replaceability of different layers. Instead of manual or heuristic choices, we frame layer selection as a binary classification problem and propose a selective transfer, which automatically and progressively converts layers to linear attention with minimal performance impact. Additionally, to overcome the ineffectiveness and even inefficiency of existing objectives in optimizing this challenge transfer process, we introduce an anytime distribution matching (ADM) objective that aligns the distributions of samples across any timestep along the sampling trajectory. This objective is highly efficient and recovers model performance. Extensive experiments show that \textsc{LinVideo} achieves a $\mathbf{1.43\text{--}1.71\times}$ speedup while preserving generation quality, and the 4-step distilled models further reduce latency by $\mathbf{15.9\text{--}20.9\times}$ with only a minor drop in visual quality.

\end{abstract}

\section{Introduction}
\label{sec:introduction}

Recently, advances in artificial intelligence--generated content (AIGC) have yielded notable breakthroughs across text~\citep{touvron2023llamaopenefficientfoundation, deepseekai2025deepseekv3technicalreport}, image~\citep{xie2025sana, flux2024}, and video synthesis~\citep{wanteam2025wanopenadvancedlargescale, kong2025hunyuanvideosystematicframeworklarge}. 
Progress in video generative models, largely enabled by the diffusion transformer (DiT) architecture~\citep{peebles2023scalable}, has been particularly striking. 
State-of-the-art video diffusion models (DMs), including the closed-source OpenAI Sora~\citep{openai2024video} and Kling~\citep{kling2024}, as well as the open-source \texttt{Wan}~\citep{wanteam2025wanopenadvancedlargescale} and CogVideoX~\citep{yang2025cogvideoxtexttovideodiffusionmodels}, effectively capture \emph{physical consistency}, \emph{semantic scenes}, and other complex phenomena. 
Nevertheless, by introducing a temporal dimension relative to image DMs, video DMs greatly increase the sequence length $n$ to be processed (\emph{e.g.}, generating a $10s$ video often entails ${>}50K$ tokens). Consequently, the \texttt{self-attention} operator, whose cost scales quadratically with $n$ in a video DM, becomes a prohibitive bottleneck for deployment.

Prior work has addressed this challenge by designing more efficient attention mechanisms. These methods fall into two categories: (\emph{i}) attention sparsification~\cite{xi2025sparsevideogenacceleratingvideo, li2025radialattentiononlogn, zhang2025fastvideogenerationsliding}, which skips redundant dense-attention computations; and (\emph{ii}) linear attention~\cite{chen2025sanavideoefficientvideogeneration} and its variants~\cite{wang2025lingenhighresolutionminutelengthtexttovideo, dalal2025oneminutevideogenerationtesttime}, which modify computation and architectures to reduce the time and memory complexity from $\mathcal{O}(n^2)$ to $\mathcal{O}(n)$. However, sparsification often cannot reach high sparsity at moderate sequence lengths and, in practice, still retains more than $50\%$ computation of quadratic dense attention. While linear attention offers much lower complexity, replacing \textit{all} quadratic attention layers with linear attention recovers video quality only through time- and resource-intensive pretraining~\cite{chen2025sanavideoefficientvideogeneration, wang2025lingenhighresolutionminutelengthtexttovideo}. This arises from (\textit{i}) the clear representation gap~\cite{zhang2024the} between quadratic and linear attention and (\textit{ii}) the complexity of spatiotemporal modeling for video generation, both of which make cost-effective post-training impractical and limit the adoption of linear attention. This paper therefore asks: \textit{Can we, via efficient post-training, replace as many quadratic attention layers as possible with linear attention so that effective inference acceleration is achieved without degrading the performance of video DMs?}

To address this challenge, we present \textsc{LinVideo}, an efficient \emph{data-free} post-training framework for a pre-trained video DM that selectively replaces a large fraction of $\mathcal{O}(n^2)$ \texttt{self-attention} with $\mathcal{O}(n)$ attention, while preserving output quality. To begin with, we construct training data from the pre-trained model’s own inputs and outputs, removing the need for curated high-quality video datasets. Then, we introduce two techniques that (\textit{i}) choose which attention layers to replace with minimal performance loss and (\textit{ii}) optimize post-training to recover the original performance, respectively.

Specifically, we find that replacing different layers with linear attention results in significantly varying performance. Guided by this observation, we propose a learning-based \textit{selective transfer} method that progressively and automatically replaces a target number of quadratic attention layers with linear attention, while incurring minimal performance loss. For each layer, we cast the choice as a binary classification problem and use a learnable scalar to produce a classification score in $[0,1]$ for the two classes (quadratic \emph{vs.} linear). After training, we round the score to pick the type of attention in inference. We also add a constraint loss to steer the total number of selected linear layers toward the target, as well as a regularization term that drives the scores toward $0/1$ to reduce rounding error and training noise.

Nevertheless, it is still challenging to optimize video DMs in the above transfer process. Direct output matching on the training set introduces temporal artifacts and weakens generalization. Few-step distillation~\cite{yin2024onestep, yin2024improved} aligns only final sample distributions and ignores intermediate timesteps, causing notable drops in our setting. Furthermore, it needs an auxiliary diffusion model to estimate the generator’s score function~\cite{song2021scorebased}. Based on these findings, we propose an \textit{anytime distribution matching} (ADM) objective that aligns sample distributions at any timestep along the sampling trajectory, and we estimate the score function using the current model itself. This objective effectively preserves model performance while enabling efficient training.

To summarize, our contributions are as follows:
\begin{itemize}[leftmargin=*, nosep]
    \item We introduce, to our knowledge, the \textit{first} efficient \textit{data-free} post-training framework, \textsc{LinVideo}, that replaces quadratic attention with linear attention in a pre-trained video DM, enabling efficient video generation without compromising performance.
    \item We propose \textit{selective transfer}, which automatically and smoothly replaces a target number of quadratic attention with linear attention, minimizing performance drop.
    \item We present an \textit{anytime distribution matching} (ADM) objective that effectively and efficiently aligns the distributions of samples from the trained model and the original DM at any timestep.
    \item Extensive experiments show that our method achieves a $\mathbf{1.43\text{--}1.71\times}$ latency speedup and outperforms prior post-training methods on VBench. Moreover, we are the \textit{first} to apply few-step distillation to a linear-attention video DM. Remarkably, the 4-step models attain a $\mathbf{15.9\text{--}20.9\times}$ speedup.
\end{itemize}

\section{Related Work}
\label{sec:related_work}

\noindent\textbf{Video DMs.} Video generation has emerged as a rapidly growing area in generative AI, with most approaches built on the denoising diffusion paradigm. A major breakthrough—enabling high compression rates and long-form generation—arrived with Sora~\cite{openai2024video}, which introduced a temporal variational auto-encoder (VAE) that compresses temporal as well as spatial dimensions and scaled up the diffusion transformer (DiT)~\cite{peebles2023scalable} architecture for video generation. Subsequent efforts~\cite{wanteam2025wanopenadvancedlargescale,seawead2025seaweed,yang2025cogvideoxtexttovideodiffusionmodels,deepmind_veo3_2025,kong2025hunyuanvideosystematicframeworklarge,gao2025wan} have further advanced this modern design space. For example, CogVideoX~\cite{yang2025cogvideoxtexttovideodiffusionmodels} introduced expert-adaptive \texttt{LayerNorm} to improve text–video fusion, while \wan2.2~\cite{wanteam2025wanopenadvancedlargescale} incorporated a sparse mixture-of-expert (MoE)~\cite{shazeer2017outrageouslylargeneuralnetworks} that routes diffusion steps to specialized experts. Alongside methodological progress, recent large-scale deployments underscore the transformative potential of video generation. Systems such as Kling~\cite{kling2024} and Seaweed~\cite{seawead2025seaweed} demonstrate substantial practical impact across creative and industrial applications. Taken together, these advances establish video generation as one of the most dynamic and competitive frontiers within the generative AI community.

\noindent\textbf{Efficient attention for video DMs.} There have been lots of studies~\cite{ding2025efficientvditefficientvideodiffusion, xi2025sparsevideogenacceleratingvideo} on video diffusion models (DMs) aiming to accelerate inference~\cite{huang2025harmonicaharmonizingtraininginference} by sparsifying computationally expensive dense 3D attention. These methods fall into two categories: \textit{static} and \textit{dynamic}. \textit{Static} methods~\cite{ sun2025vortaefficientvideodiffusion, zhang2025fastvideogenerationsliding, zhang2025vsafastervideodiffusion, li2025radialattentiononlogn} predefine a sparsity pattern offline by identifying critical tokens. However, these methods fail to capture the dynamics of sparsity patterns during inference, leading to suboptimal performance. \textit{Dynamic} methods~\cite{zhang2025spargeattentionaccuratetrainingfreesparse, xu2025xattentionblocksparseattention, tan2025dsvexploitingdynamicsparsity, zhang2025trainingfreeefficientvideogeneration, xi2025sparsevideogenacceleratingvideo,lv2026lightforcingacceleratingautoregressive} adapt the sparsity pattern at inference time based on input content. Thus, they need to select critical tokens through an additional identification step. Other approaches~\cite{zhao2025paroattentionpatternawarereorderingefficient, liu2025fpsattentiontrainingawarefp8sparsity} combine sparsification with quantization~\cite{huang2024tfmqdmtemporalfeaturemaintenance,huang2025temporalfeaturemattersframework,huang2026qvgenpushinglimitquantized} to further reduce latency and memory usage.

Meanwhile, linear attention~\cite{esser2020learnedstepsizequantization} and its alternatives (\emph{e.g.}, state space models~\cite{gu2024mambalineartimesequencemodeling}) for video generation have also gained attention. Most of these works~\cite{gao2024mattenvideogenerationmambaattention, wang2025lingenhighresolutionminutelengthtexttovideo, huang2025m4vmultimodalmambatexttovideo, dalal2025oneminutevideogenerationtesttime, chen2025sanavideoefficientvideogeneration} focus on a costly pretraining manner initialized from a given image generative model~\cite{xie2024sana}. Among them, Matten~\cite{gao2024mattenvideogenerationmambaattention} and LinGen~\cite{wang2025lingenhighresolutionminutelengthtexttovideo} adopt Mamba~\cite{gu2024mambalineartimesequencemodeling} to capture global information. SANA-Video~\cite{chen2025sanavideoefficientvideogeneration} involves auto-regressive training~\cite{chen2024diffusion} with block-wise causal linear attention. To the contrary, we explore applying linear attention to advanced pre-trained video DMs~\cite{wanteam2025wanopenadvancedlargescale}. One concurrent work, SLA~\cite{zhang2025slasparsitydiffusiontransformers}, proposes intra-layer mixed attention (\emph{e.g.}, quadratic and linear attention) in a post-training manner. Instead, we concentrate on inter-layer mixed attention (\emph{i.e.}, replacing partial layers with linear attention). Moreover, we believe our data-free finetuning with the proposed strategy can be combined with SLA for more efficient and high-performing linearization for video DMs.

\section{Preliminaries}
\label{sec:preliminaries}
\noindent\textbf{Video diffusion modeling.} The video DM~\citep{ho2022video, opensora} extends image DMs~\citep{li2023qdm, Song2021DenoisingDI} into the temporal domain by learning dynamic inter-frame dependencies. Let $\bx_0 \in \mathbb{R}^{f\times h\times w\times c}$ be a latent video variable, where $f$ denotes the count of video frames, each of size $h\times w$ and with $c$ channels. DMs are trained to denoise samples generated by adding random Gaussian noise $\bepsilon\sim\mathcal{N}(\mathbf{0}, \mathbf{I})$ to $\bx_0$:
\begin{equation}
        \bx_t=\alpha_t\bx_0+\sigma_t\bepsilon,
    \label{eq:add_noise}
\end{equation}
where $\alpha_t\ge 0,\sigma_t>0$ are specified noise schedules such that $\frac{\alpha_t}{\sigma_t}$ is monotonically decreasing \emph{w.r.t.} timestep $t$ and a larger $t$ indicates greater noise. With noise prediction parameterization~\citep{ho2020denoisingdiffusionprobabilisticmodels} and discrete-time schedules (\emph{i.e.}, $t\in[1,\ldots, T]$ and typically $T=1000$), the training objective for a neural network $\bepsilon_\theta$ parameterized by $\theta$ can be formulated as follows:
\begin{equation}
        \mathbb{E}_{\bx_0, \bepsilon, \mathcal{C}, t}\!\left[w(t)\|\bepsilon-\bepsilon_\theta(\bx_t, \mathcal{C}, t)\|^2_F\right],
\end{equation}
where $\mathcal{C}$ represents conditional guidance, such as texts or images, $w(t)$ is a weighting function, and $\|\cdot\|_F$ denotes the Frobenius norm. Advanced video DMs~\cite{wanteam2025wanopenadvancedlargescale, kong2025hunyuanvideosystematicframeworklarge} are \textit{flow matching models}~\cite{liu2022flow}. They employ velocity prediction parameterization~\cite{salimans2022progressive} and continuous-time coefficients. In \textit{rectified flow models}~\cite{wang2024rectified}, the standard choice is $\alpha_t = 1 - t, \sigma_t = t$ for $t \in [0, 1]$, which is the setting adopted in this work. The conditional probability path or the velocity is given by $\bm{v}_t = \frac{d\alpha_t}{dt} \bx_0 + \frac{d\sigma_t}{dt} \bepsilon$, and the corresponding training objective is:
\begin{equation}
\mathbb{E}_{\bx_0, t, \mathcal{C}, \bepsilon}\!\left[w(t) \| \bm{v}_t - \bm{v}_\theta(\bx_t,\mathcal{C}, t) \|_F^2\right],
\label{eq:v_loss}
\end{equation}
where $\bm{v}_\theta$ is a neural network parameterized by $\theta$. The sampling procedure of these models begins at $t = 1$ with $\bx_1 \sim \mathcal{N}(\bm{0}, \bm{I})$ and stops at $t = 0$, solving the \textit{Probability-Flow  Ordinary Differential Equation} (PF-ODE) by $d\bx_t = \bm{v}_\theta(\bx_t, \mathcal{C}, t) dt$.

After obtaining $\bx_0$ through iterative sampling, the raw video is obtained by decoding the variable via a video variational auto-encoder (VAE)~\cite{wanteam2025wanopenadvancedlargescale}.

\noindent\textbf{Attention computation.} Given input $\vx\in\sR^{n\times d}$ (where $n$ denotes the sequence length and $d$ signifies the feature dimension), attention is written as:
\begin{equation}
    \vo_i=\sum^n_{j=1}\frac{\texttt{sim}(\vq_i, \vk_j)}{\sum^n_{j=1}\texttt{sim}(\vq_i, \vk_j)}\vv_j,
    \label{eq:attention}
\end{equation}
where $\vq=\vx\mW_q, \vk=\vx\mW_k, \vv=\vx\mW_v$ and $\mW_q/\mW_k/\mW_v\in\sR^{d\times d}$ are learnable projection matrices. $i/j$ are row indices for their corresponding matrices and $\texttt{sim}(\cdot, \cdot)$ indicates the similarity function. When employing $\texttt{sim}(\vq, \vk)=\exp(\frac{\vq\vk^\top}{\sqrt{d}})$, Eq.~(\ref{eq:attention}) represents standard \textit{softmax attention}~\cite{vaswani2023attentionneed}. In this way, the attention map computes the similarity between all query-key pairs, which causes the computational complexity to be $\mathcal{O}(n^2)$. 

In contrast, \textit{linear attention}~\cite{katharopoulos2020transformersrnnsfastautoregressive} adopts a carefully designed kernel $k(x,y)=\langle \phi(x), \phi(y)\rangle$  as the approximation of the original function (\emph{i.e.}, $\texttt{sim}(\vq, \vk)=\phi(\vq)\phi(\vk)^\top$). In this case, we can leverage the associative property of matrix multiplication to reduce the computational complexity from $\mathcal{O}(n^2)$ to $\mathcal{O}(n)$ without changing functionality:
\begin{equation}
    \vo_i=\sum^n_{j=1}\frac{\phi(\vq_i)\phi(\vk_j)^\top}{\sum^n_{j=1}\phi(\vq_i)\phi(\vk_j)^\top}\vv_j = \frac{\phi(\vq_i)\big(\sum^n_{j=1}\phi(\vk_j)^\top\vv_j\big)}{\phi(\vq_i)\big(\sum^n_{j=1}\phi(\vk_j)^\top\big)}.
    \label{eq:linear_attention}
\end{equation}
In this work, we directly employ the effective kernel design from \textit{Hedgehog}~\cite{zhang2024the}\footnote{Ablation study for kernel function can be found in Sec.~\ref{sec:kernel}.}, which is formulated as:
\begin{equation}
\phi(\vq)=\texttt{softmax}(\vq\widetilde\mW_q)\oplus\texttt{softmax}(-\vq\widetilde\mW_q),
\label{eq:kernel}
\end{equation}
where $\widetilde\mW_q\in \sR^{d\times\frac{d}{2}}$ is a learnable matrix. Both $\oplus$ (concat) and $\texttt{softmax}(\cdot)$ apply to the feature dimension. The same function applies to $\vk$ with another learnable matrix $\widetilde\mW_k\in \sR^{d\times\frac{d}{2}}$.

\section{\textsc{LinVideo}}\label{sec:method}
\begin{figure*}[ht]
\vspace{-0.3in}
\centering
\setlength{\abovecaptionskip}{0.2cm}
\includegraphics[width=0.9\textwidth]{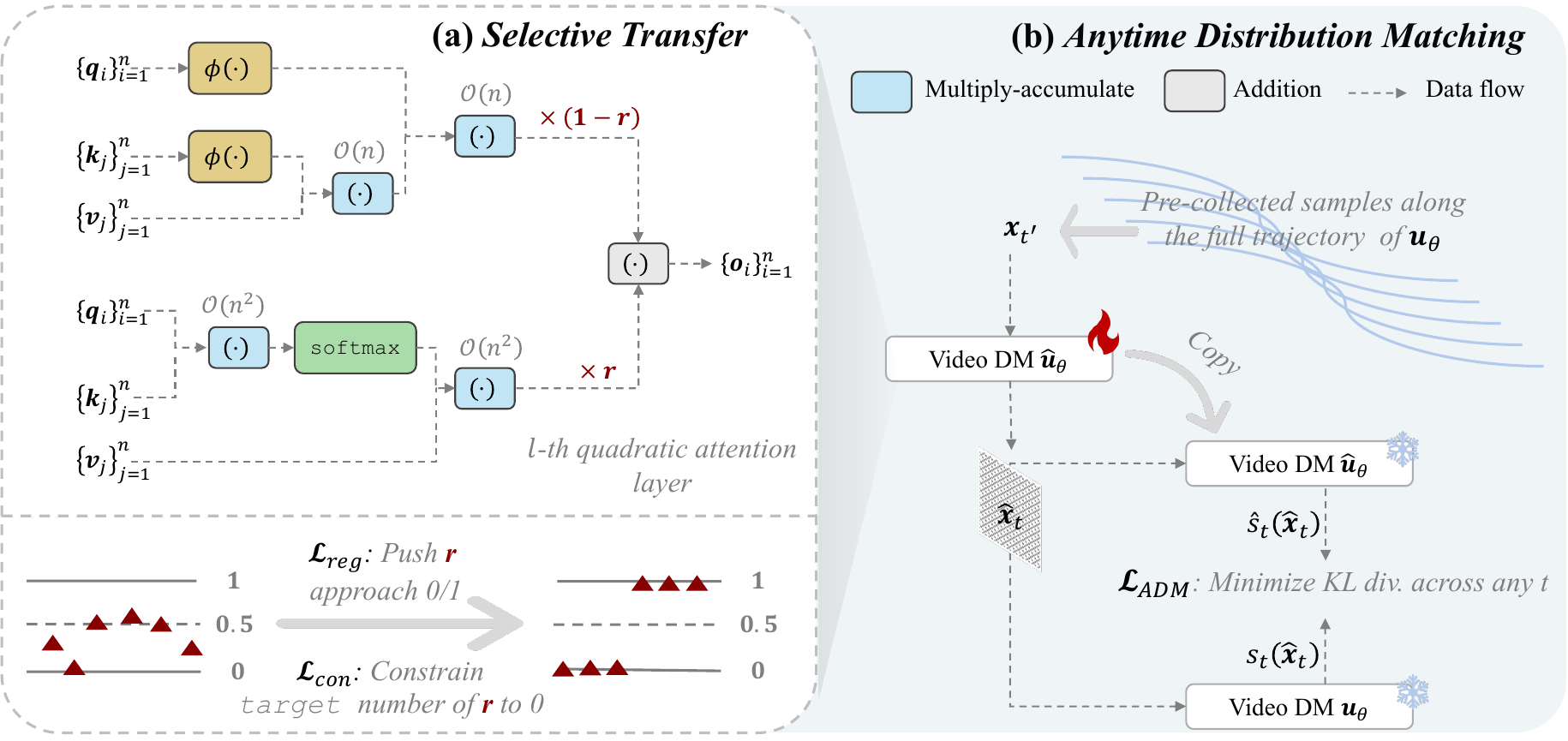}
\vspace{-0.05in}
\caption{Overview of the proposed efficient \textit{data-free} post-training framework, \textsc{LinVideo}. (a) This framework first applies \textit{selective transfer} (Sec.~\ref{sec:select_transfer}), which assigns each layer a learnable score $r$ and progressively, automatically replaces quadratic attention with linear attention while minimizing the resulting performance drop. This process also combines with $\mathcal{L}_\text{con}$ (\emph{i.e.}, Eq.~(\ref{eq:constraint})) and $\mathcal{L}_\text{reg}$ (\emph{i.e.}, Eq.~(\ref{eq:reg})) to ensure a given \texttt{target} number of layers replaced by linear attention and mitigate the fluctuation (around $0.5$) of $r$ to improve training, respectively. (b) Moreover, \textsc{LinVideo} integrates an \textit{anytime distribution matching} objective (Sec.~\ref{sec:tdm}), which aims to match the sample distributions between $\hat{\vu}_\theta$ and $\vu_\theta$ across any timestep in the sampling trajectory. This significantly recovers performance and enables high efficiency compared with previous objectives in our scenarios.}
\label{fig:overview}
\vspace{-0.2in}
\end{figure*}

In this work, we propose \textsc{LinVideo} (see Fig.~\ref{fig:overview} ), a post-training framework that selectively replaces softmax attention with linear attention for a pre-trained video DM.

\noindent\textbf{Preparation for \textit{data-free} post-training.} Video diffusion models~\cite{wanteam2025wanopenadvancedlargescale, yang2025cogvideoxtexttovideodiffusionmodels, openai2024video} demand large, diverse datasets, yet access is often limited by scale, privacy, and copyright. To this end, we adopt a \textit{data-free fine-tuning} approach in this work, which transfers the original model’s (\emph{i.e.}, $\vu_\theta$) prediction ability to its linear attention version without requiring the original dataset. Specifically, we first randomly sample a large amount of initial noise $\bx_1 \sim \mathcal{N}(\bm{0}, \bm{I})$, and fetch all the input and output pairs of $\vu_\theta$ in the sampling trajectory (start from $\bx_1$) (\emph{i.e.}, $(\bx_t, \vu_t)$\footnote{We omit the conditions for simplicity.} for $t\in[0,1]$) as our training dataset and targets. During fine-tuning, a naive objective can be defined as:
\begin{equation}
    \mathcal{L}_\text{mse}=\|\vu_t-\hat\vu_\theta(\bx_t, t)\|_F^2,
    \label{eq:naive_objective}
\end{equation}
where $\hat\vu_\theta$ is our video DM with linear attention. However, due to the significant representation capability gap~\cite{zhang2024the, katharopoulos2020transformersrnnsfastautoregressive} between the softmax attention and linear attention, it is challenging to preserve the complex temporal and spatial modeling capabilities of video DMs when (\textit{i}) directly replacing \textit{all} quadratic attention layers with linear attention modules and (\textit{ii}) employing naive \textit{data-free fine-tuning}. 

In light of this, we propose two novel techniques, \emph{i.e.}, \textit{selective transfer} (Sec.~\ref{sec:select_transfer}) and \textit{anytime distribution matching}  (Sec.~\ref{sec:tdm}) to replace maximum softmax attention layers with linear attention while preserving the video DM's exceptional performance. The training overview is presented in Sec.~\ref{sec:overview}.

\subsection{Selective Transfer for Effective Linearization}\label{sec:select_transfer}
In this subsection, we explore how to effectively select layers for linear attention replacement.

\noindent\textbf{Layer selection matters.} As described before, we consider replacing partial layers in this work, and we first find that the choice of which layers to replace in a pre-trained video DM can lead to significant performance disparities after fine-tuning. As shown in Fig.~\ref{fig:disparity}, this disparity follows two main patterns: 
\begin{figure}[tp!]
   \centering
    \setlength{\abovecaptionskip}{0.2cm}
        \includegraphics[width=0.48\textwidth]{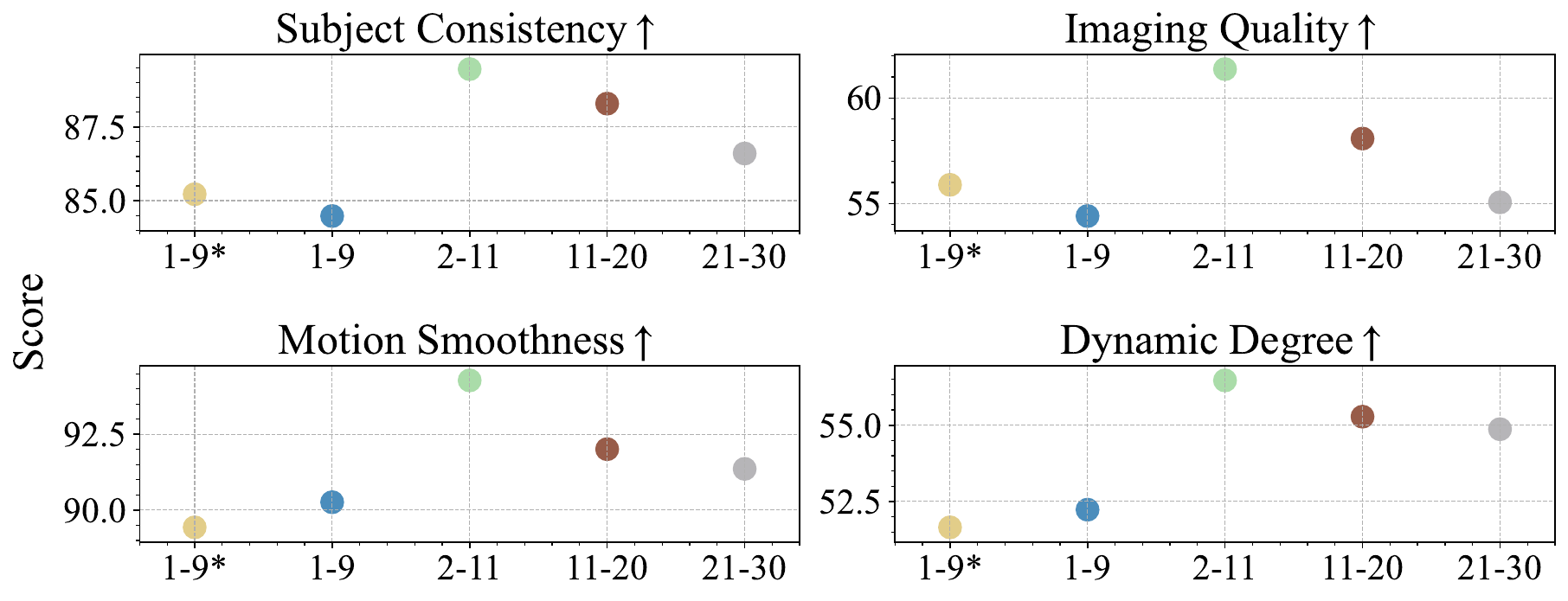}
        \vspace{-0.2in}
        \caption{\label{fig:disparity} Performance on $4$ VBench~\cite{huang2024vbench} dimensions for partial linearized ($10$ adjacent layers for each dot) \wan $1.3$B~\cite{wanteam2025wanopenadvancedlargescale} after $2K$-step fine-tuning. The index range of the layers replaced with linear attention is indicated in the tick label of the x-axis. ``*'' denotes models further fine-tuned for $3K$ additional steps.}
        \vspace{-0.15in}
\end{figure}

\noindent(\textit{i}) \uline{Models with linearized shallow layers recover accuracy more easily than models with linearized deep layers.} For instance, applying linear attention from the $2$-nd to the $11$-th layer achieves improvements of $+2.86$ and $+6.31$ in Subject Consistency and Image Quality, respectively, compared to applying it from the $21$-th to the $30$-th layer. This may be because errors introduced by shallow layers can be better compensated by the optimization of subsequent layers. 

\noindent(\textit{ii}) \uline{However, including certain layers, such as the first layer (\emph{i.e.}, \bl{blue} dots), results in significant performance drops when replaced}. These declines can not be mitigated after extended fine-tuning (\emph{i.e.}, \yl{yellow} dots).

\noindent\textbf{Selective transfer from $\mathcal{O}(n^2)$ to $\mathcal{O}(n)$.} Based on the above investigation, we propose \textit{selective transfer} to select partial quadratic attention layers with linear attention replacement. This approach automatically determines which layers to replace.
It also enables a progressive and smooth transition from the original softmax attention to linear attention. Specifically, inspired by the binary classification problem, we consider each type of attention (for each layer) as an individual class, opposite to the other. Then, we employ a mixed-attention computation as:

\begin{equation}
   \vo_i=r\sum^n_{j=1}\frac{\texttt{exp}(\frac{\vq_i\vk_j^\top}{d})}{\sum^n_{j=1}\texttt{exp}(\frac{\vq_i\vk_j^\top}{d})}\vv_j +  (1-r)\frac{\phi(\vq_i)\big(\sum^n_{j=1}\phi(\vk_j)^\top\vv_j\big)}{\phi(\vq_i)\big(\sum^n_{j=1}\phi(\vk_j)^\top\big)},
    \label{eq:training_attention}
\end{equation}

where $\phi(\cdot)$ follows Eq.~(\ref{eq:kernel}) and $r$ is an introduced learnable parameter\footnote{In detail, each $r$ is clipped into $[0, 1]$ before computing.}. In Eq.~(\ref{eq:training_attention}), $r$ and $1-r$ represent the classification score for each class (\emph{i.e.}, quadratic and linear attention, respectively). We initialize $[r^{(1)},\ldots, r^{(N)}]=\bI\in\sR^{1\times N}$ to stabilize the training, where $N$ denotes the maximum index of the attention layers. After training, if the classification score for a class is greater than $0.5$, the corresponding class is selected for inference. This means we preserve the quadratic attention and remove the linear attention branch when $\lceil r\rfloor=1$, and \emph{vice versa}.

Here, we propose a constraint loss to enforce the video DM with $\texttt{target}$ (given before training) layers being replaced with linear attention:
\begin{equation}
    \mathcal{L}_\text{con}=\big(\sum^N_{l=1}\lceil r^{(l)}\rfloor - \texttt{target}\big)^2.
    \label{eq:constraint}
\end{equation}
To ensure the differentiability, we employ a straight-through estimator (STE)~\citep{bengio2013estimatingpropagatinggradientsstochastic} to $r$ as $\frac{\partial \lceil r\rfloor}{\partial r}=1$.

\begin{figure}[ht!]
\vspace{-0.1in}
   \centering
    \setlength{\abovecaptionskip}{0.2cm}
        \includegraphics[width=0.47\textwidth]{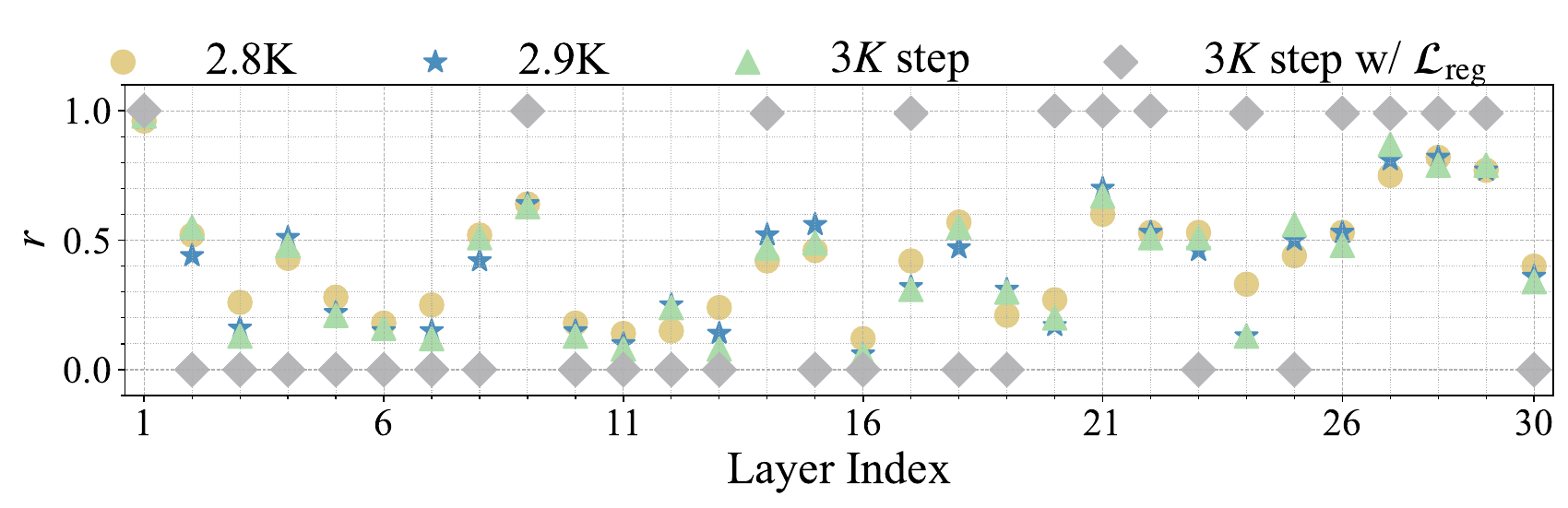}
        \vspace{-0.1in}
        \caption{\label{fig:rounding} Values of $r$ across layers and training steps. ``w/ $\mathcal{L}_\text{reg}$'' denotes we employ Eq.~(\ref{eq:reg}) for training, otherwise only Eq.~(\ref{eq:constraint}) is applied to guide the training of $r$.}
        \vspace{-0.1in}
\end{figure}
Moreover, to force $r$ to approach $0/1$ during training, inspired by model quantization~\cite{nagel2020downadaptiveroundingposttraining}, we further apply a regularization: 
\begin{equation}
    \mathcal{L}_\text{reg}=\sum_{i=1}^N\big(1-|2r^{(l)}-1|^\alpha\big),
    \label{eq:reg}
\end{equation}
where we annealing decay the parameter $\alpha$ from large to small. This encourages $r$ to move more adaptively at the initial phase to improve the training loss, but forces it to $0/1$ in the later phase (see this effect in the Appendix). This regularization helps mitigate the significant performance degradation caused by $\lceil\cdot \rfloor$. As demonstrated in Fig.~\ref{fig:rounding}, without $\mathcal{L}_\text{reg}$,  a large number of $r$ values fluctuate around the $0.5$ boundary at the end of training. As both attentions in Eq.~(\ref{eq:training_attention}) occupy a non-negligible proportion (\emph{i.e.}, $r\approx 0.5$), directly removing one of them by rounding $r$ for inference can result in significant accuracy drops (as shown in Sec.~\ref{sec:ablation}). Moreover, the fluctuation of $r$ introduces training noise, which could also disrupt the learning process. 

\subsection{Match the Distribution across Any Timestep}\label{sec:tdm}
Our \textit{selective transfer} determines how to select layers for linear attention replacement. In this subsection, we study how to optimize this process. 

\noindent\textbf{Objective analysis.} The naive optimization objective (\emph{i.e.}, Eq.~(\ref{eq:naive_objective})) leads to temporal artifacts (\emph{e.g.}, flicker and jitter), as demonstrated in the Appendix, likely because it does not preserve the model’s original joint distribution over frames. This objective also harms generalization by enforcing exact latent alignment on the training set~\cite{yin2024onestep}. Prior few-step distillation works~\cite{yin2024improved, yin2024slow, yin2025causvid, ge2025senseflowscalingdistributionmatching} address above problems using \textit{distribution matching}, which seeks to align $p_g$\footnote{The distribution of outputs $\bx_g$, which is generated by a few-step generator.} with $p_0$\footnote{The distribution for the final sample $\bx_0$, which is generated by the original DM.} by minimizing their Kullback–Leibler (KL) divergence during training. However, directly applying this objective in our work faces two key challenges: 

\noindent(\textit{i}) \uline{\textit{Distribution matching} in few-step distillation only matches $p_g$ against $p_0$, discarding distribution $p_t$}\footnote{$p_t$ is the distribution of $\bx_t$, which is the sample generated by the original DM at $t$.} \uline{across timesteps.} This leads to considerable performance degradation in our scenario, as demonstrated in Sec.~\ref{sec:ablation}.

\noindent(\textit{ii}) \uline{An additional multi-step DM must be trained to approximate the score~\cite{song2021scorebased} determined by $p_g$, which is required to compute the gradient of the KL divergence between $p_g$ and $p_0$.} This necessity arises because the few-step generator learns only the naive mapping: $p_T\mapsto p_0$, while its score field remains intractable~\cite{yin2024improved, yin2024onestep}. Even worse, the additional multi-step DM  typically incurs $5\text{--}10\times$ generator’s training cost, rendering the approach inefficient.

\begin{table*}[ht!]\setlength{\tabcolsep}{1pt}
 \renewcommand{\arraystretch}{1.05}
  \centering
   \vspace{-0.3in}
  \caption{Performance comparison with relevant baselines on $8$ dimensions of VBench~\cite{huang2024vbench}. ``+DMD2'' denotes our 4-step distilled \textsc{LinVideo} model. We highlight the best score and the second score in \textbf{bold} and \underline{underlined} formats, respectively. More results can be found in Sec.~\ref{sec:comp-few} and Sec.~\ref{sec:cog}.} 
  \vspace{-0.1in}
  \resizebox{0.95\linewidth}{!}{

\begin{tabu}[t!]{l|cc|cccccccc}
\toprule
\multirow{2}{*}{Method}  & \multirow{2}{*}{Latency~(s)$\downarrow$} & \multirow{2}{*}{Speedup$\uparrow$} & \multirow{2}{*}{\rotatebox{0}{\makecell{Imaging\\Quality}$\uparrow$}} &  \multirow{2}{*}{\rotatebox{0}{\makecell{Aesthetic\\Quality}$\uparrow$}} & \multirow{2}{*}{\rotatebox{0}{\makecell{Motion\\Smoothness}$\uparrow$}} & \multirow{2}{*}{\rotatebox{0}{\makecell{Dynamic \\Degree}$\uparrow$}} & \multirow{2}{*}{\rotatebox{0}{\makecell{Background\\Consistency}$\uparrow$}}  & \multirow{2}{*}{\rotatebox{0}{\makecell{Subject \\Consistency}$\uparrow$}} & \multirow{2}{*}{\rotatebox{0}{\makecell{Scene\\Consistency}$\uparrow$}} & \multirow{2}{*}{\rotatebox{0}{\makecell{Overall\\Consistency}$\uparrow$}}\\
 & & & & & & \\
\midrule
\multicolumn{11}{c}{\cellcolor[gray]{0.92}\wan 1.3B ($\texttt{CFG}=5.0, 480p, \texttt{fps}=16$)} \\
\midrule
 FlashAttention2~\cite{dao2023flashattention2fasterattentionbetter}  & 97.32  & $1.00\times$ & 66.25 &59.49 &98.42 &59.72 &96.57 &95.28 &39.14 &26.18  \\
 \midrule
 DFA~\cite{yuan2024ditfastattnattentioncompressiondiffusion} & 88.95 & $1.09\times$ &  65.41 & 58.35 & \underline{98.11} & 58.47 & 95.82 & 94.31 &  38.43 & 26.08\\
 XAttn~\cite{xu2025xattentionblocksparseattention} &  83.51 & $1.17\times$ &65.32 & 58.51 & 97.42 & 59.02 & 95.43 & 93.65 & 38.14 & 26.22 \\
 SVG~\cite{xi2025sparsevideogenacceleratingvideo} &  74.52 & $1.31\times$ & 65.78 & 59.16 & 97.32 & 58.87 & 95.79 & 93.94 & 38.54 & 25.87\\
 SVG2~\cite{yang2025sparsevideogen2acceleratevideo} &  84.91 & $1.15\times$ &\underline{66.03} & \underline{59.31} & 98.07 & 59.44 & \underline{96.61} & \underline{94.95} & \underline{39.14} & \underline{26.48}\\
\midrule
\rowcolor{mycolor!30}\textsc{LinVideo} &  \underline{68.26} & \underline{$1.43\times$} & \textbf{66.07} & \textbf{59.41} & \textbf{98.19} & \underline{59.67} & \textbf{96.72} & \textbf{95.12} & \textbf{39.18} & \textbf{26.52}\\
\rowcolor{mycolor!30}\textsc{LinVideo} + DMD2~\cite{yin2024improved}  & \textbf{6.110} & $\mathbf{15.9\times}$ & 65.62 & 57.74 & 97.32 & \textbf{61.26} & 95.47 & 93.74 & 38.78 & 25.94\\

\midrule
\multicolumn{11}{c}{\cellcolor[gray]{0.92}\wan 14B ($\texttt{CFG}=5.0, 720p, \texttt{fps}=16$)} \\
\midrule
 FlashAttention2~\cite{dao2023flashattention2fasterattentionbetter}& 1931 & $1.00\times$ & 67.89 &61.54 &97.32 &70.56 &96.31 &94.08 &33.91 &26.17  \\
 \midrule
 DFA~\cite{yuan2024ditfastattnattentioncompressiondiffusion} & 1382 & $1.40\times$ & 65.93 & 60.13 & 96.87 & 69.34 & 95.37 & 93.26 & 33.14 & 26.12 \\
 XAttn~\cite{xu2025xattentionblocksparseattention} & 1279 & $1.51\times$ & 65.47 & 60.36 & 96.28 & 69.25 & 95.24 & 92.97 & 33.22 & \underline{26.14} \\
 SVG~\cite{xi2025sparsevideogenacceleratingvideo} & 1203 & $1.61\times$ & 66.09 & 60.86 & 96.91 & 69.46 & 95.35 & 93.18 & 33.46 & 26.07\\
 SVG2~\cite{yang2025sparsevideogen2acceleratevideo} & 1364 & $1.42\times$ & \underline{66.25} & \underline{61.08} & \underline{97.12} & 69.43 & \underline{95.51} & \underline{93.39} & \underline{33.52} & \underline{26.14} \\
\midrule
\rowcolor{mycolor!30}\textsc{LinVideo} & \underline{1127} & \underline{$1.71\times$} & \textbf{66.47} & \textbf{61.36} & \textbf{97.24} & \textbf{69.82} & \textbf{96.34} & \textbf{93.68} & \textbf{33.72} & \textbf{26.16}\\
\rowcolor{mycolor!30}\textsc{LinVideo} + DMD2~\cite{yin2024improved} & \textbf{92.56} & $\mathbf{20.9\times}$ & 65.74 & 59.68 & 96.32 & \underline{69.74} & 95.38 & 92.88 & 33.18 & 26.09\\

\bottomrule
\end{tabu}
}
    \label{tab:compare}
    \vspace{-0.1in}
\end{table*}

\noindent\textbf{Anytime distribution matching (ADM).}
To this end, we propose an \textit{anytime distribution matching} (ADM) method to address challenge (\textit{i}). Instead of just matching the final data distributions ($t=0$), the core idea is to match the distributions across any timestep $t\in[0,1]$ along the entire sampling trajectory. This objective encourages the linearized DM to produce samples whose distribution at every $t$ matches that of the original DM. Specifically, let $q_t$ denote the distributions of $\hat{\bx}_t$\footnote{Samples generated by the linearized DM.}, respectively.
For any given $t$, we minimize the KL divergence between $q_t$ and $p_t$:
\begin{equation}
    \begin{split}
        \mathcal{L}_{\mathrm{ADM}}
        &= \mathbb{E}_{\hat{\bx}_t \sim q_t}\!\left[\log\frac{q_t(\hat{\bx}_t)}{p_t(\hat{\bx}_t)}\right] \\
        &= \mathbb{E}_{\hat{\bx}_t \sim q_t}\!\left[-\big(\log p_t(\hat{\bx}_t) - \log q_t(\hat{\bx}_t)\big)\right].
    \end{split}
    \label{eq:kl}
\end{equation}
Here, $\hat{\bx}_t = (t-t')\,\hat{\vu}_\theta(\bx_{t'}, t') + \bx_{t'}$\footnote{We employ the Euler solver for simplicity.},
where $t$ and $t'$ are adjacent timesteps on the sampling trajectory, and $\bx_{t'}$ is collected sample from the original DM in preparation.
The gradient of Eq.~(\ref{eq:kl}) with respect to the parameters $\theta$ of $\hat{\vu}_\theta$ can be written as:
\begin{equation}
    \frac{\partial \mathcal{L}_{\mathrm{ADM}}}{\partial \theta}
    = \mathbb{E}_{\hat{\bx}_t \sim q_t}\!\left[-\big(s_t(\hat{\bx}_t) - \hat{s}_t(\hat{\bx}_t)\big)\,
    \frac{\partial \hat{\bx}_t}{\partial \hat{\vu}_\theta}\,
    \frac{\partial \hat{\vu}_\theta}{\partial \theta}\right],
    \label{eq:expand_kl}
\end{equation}
where $s_t(\hat{\bx}_t)=\nabla_{\hat{\bx}_t}\log p_t(\hat{\bx}_t), \hat{s}_t(\hat{\bx}_t)=\nabla_{\hat{\bx}_t}\log q_t(\hat{\bx}_t)$ are the score functions of $p_t$ and $q_t$, respectively.
In Eq.~(\ref{eq:expand_kl}), $s_t$ pulls $\hat{\bx}_t$ toward the modes of $p_t$, whereas $-\hat{s}_t$ pushes $\hat{\bx}_t$ away from those of $q_t$. 

Under the \textit{rectified-flow} modeling, we estimate $s_t$ with $\vu_\theta$ following Ma \emph{et al.}~\cite{ma2024sitexploringflowdiffusionbased}.
For $\hat{s}_t$, we use the model currently being trained—$\hat{\vu}_\theta$—which, as a multi-step DM, can estimate its own score function at $\hat{\bx}_t$~\cite{song2021scorebased}.
This property addresses challenge (\textit{ii}) and substantially improves training efficiency and model performance (see Sec.~\ref{sec:ablation}).
Therefore, the score difference admits the following form (see Appendix for a detailed derivation):
\begin{equation}
    s_t(\hat{\bx}_t) - \hat{s}_t(\hat{\bx}_t)
    = - \frac{1-t}{t}\,\big(\vu_\theta(\hat{\bx}_t) - \hat{\vu}_\theta(\hat{\bx}_t)\big).
    \label{eq:score}
\end{equation}

\subsection{Training Overview}~\label{sec:overview}
In summary, we first collect data pairs from the original video DM. Then we apply learnable parameters $[r^{(1)},\ldots, r^{(N)}]$ with mixed attention (\emph{i.e.}, Eq.~(\ref{eq:training_attention})) to the model. For training, we adopt the following training loss:
\begin{equation}
\mathcal{L}_\text{total}=\mathcal{L}_\text{ADM}+\lambda\big(\mathcal{L}_\text{con}+\mathcal{L}_\text{reg}\big),
    \label{eq:first-loss}
\end{equation}
where $\lambda$ is a hyper-parameter. During training, $\hat\vu_\theta$ progressively transfers from a pre-trained multi-step video DM to a high-efficiency video DM with \texttt{target} softmax attention layers replaced with linear attention (\emph{i.e.}, $r$ from $1$ to $0$).

Moreover, to further enhance inference efficiency, we provide an option to apply a few-step distillation~\cite{yin2024improved} after the above process. To be noted, directly distilling the original DM into a few-step generator with linear attention incurs catastrophic performance drops (see Appendix). 

\section{Experiments}
\label{sec:experiments}

\begin{figure*}[ht!]
\vspace{-0.3in}
   \centering
    \setlength{\abovecaptionskip}{0.2cm}
        \includegraphics[width=\textwidth]{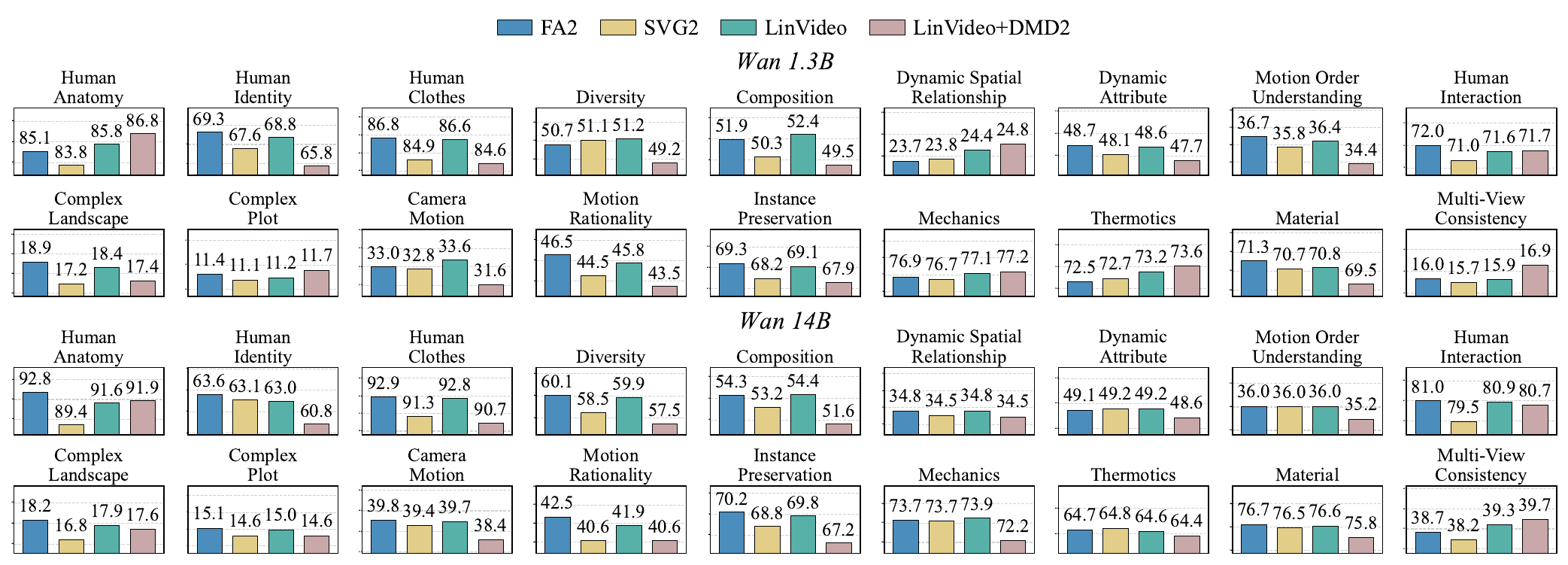}
        \vspace{-0.21in}
        \caption{Performance comparison with baselines on VBench-2.0~\cite{zheng2025vbench20advancingvideogeneration}. For \wan1.3B, the total scores are $56.74$ (FA2), $55.81$ (SVG2), $56.74$ (\textsc{LinVideo}), and $55.51$ (\textsc{LinVideo}+DMD2); for \wan14B, the total scores are $59.85$ (FA2), $58.74$ (SVG2), $59.62$ (\textsc{LinVideo}), and $58.22$ (\textsc{LinVideo}+DMD2). FA2 denotes FlashAttention2~\cite{dao2023flashattention2fasterattentionbetter}.}
        \label{fig:vbench2}
        \vspace{-0.05in}
\end{figure*}

\begin{figure*}[ht!]
    \centering
    \includegraphics[width=0.9\textwidth]{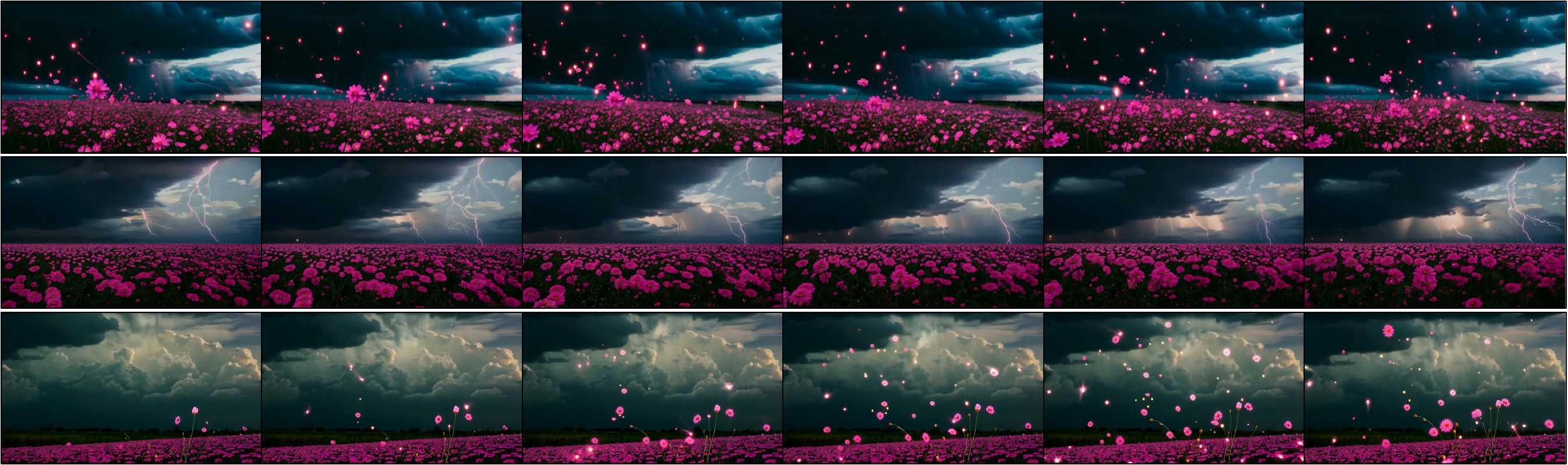}
    \vspace{-0.1in}
    \caption{Visual results at $480p$ across \wan 1.3B~\cite{wanteam2025wanopenadvancedlargescale} (\textit{Upper}), 1.3B \textsc{LinVideo} (\textit{Middle}), and 1.3B \textsc{LinVideo}+ 4-step DMD2~\cite{yin2024improved} (\textit{Lower}). Prompt: \textit{``A wide pink flower field under a stormy twilight sky with a faint magical glow. Buds burst into luminous blossoms that spread in waves across the meadow. Above, massive
black storm clouds roll hard and fast, with layered billows, shelf-cloud structure, and clear turbulence; inner lightning pulses for drama, no ground strikes.
A few bioluminescent motes drift between flowers; faint aurora-like ribbons sit behind the storm.''}}
    \label{fig:vis}
    \vspace{-0.2in}
\end{figure*}

\subsection{Experimental Details}\label{sec:experiment_details}
\noindent\textbf{Implementation.} We implement \textsc{LinVideo} with $50$-step \wan1.3B and \wan14B models~\cite{wanteam2025wanopenadvancedlargescale}, both \textit{rectified flow} text-to-video models that can generate $5s$ videos at $16$ FPS with a resolution of $832\times480$ (\textit{former}) and $720\times1280$ (\textit{latter}). Specifically, we first collect $50K$ inputs and outputs pairs from the model as the training dataset. Then, we set \texttt{target} (see Eq.~(\ref{eq:constraint})) to $16$ for \wan1.3B and $22$ for \wan 14B, which means we replace $\frac{16}{30}$ or $\frac{22}{40}$ of quadratic attention layers with linear attention. For training, the AdamW~\cite{loshchilov2019decoupledweightdecayregularization} optimizer is utilized with a weight decay of $10^{-4}$. We employ a cosine annealing schedule to adjust the learning rate over training. Additionally, we
train the models for $3K$ steps on $8\times$ H100 GPUs for \wan1.3B and $32\times$ H100 GPUs for \wan14B. For our 4-step distilled model, we employ DMD2~\cite{yin2024improved} and train the model for an additional $2K$ steps. Latency in this work is tested on a single H100 GPU. More implementation details are in the Appendix.

\noindent\textbf{Evaluation.} We select $8$ dimensions in VBench~\cite{huang2024vbench} with unaugmented prompts to comprehensively evaluate the performance following previous studies~\cite{zhao2025viditq, ren2024consisti2venhancingvisualconsistency}. Moreover, we additionally report the results on
VBench-2.0~\cite{zheng2025vbench20advancingvideogeneration} with augmented prompts to measure the adherence of videos to \textit{physical laws}, \textit{commonsense reasoning}, \emph{etc.}

\noindent\textbf{Baselines.} We compare \textsc{LinVideo} with sparse-attention baselines, including the static methods Sparse VideoGen (SVG)~\cite{xi2025sparsevideogenacceleratingvideo}, Sparse VideoGen 2 (SVG2)~\cite{yang2025sparsevideogen2acceleratevideo}, and DiTFastAttn (DFA)~\cite{yuan2024ditfastattnattentioncompressiondiffusion}, and the dynamic method XAttention~\cite{xu2025xattentionblocksparseattention}. For latency, we include only the fast attention implementations of these methods to ensure a fair comparison, excluding auxiliary designs like the \texttt{RMSNorm} kernel in SVG. To be noted, we also exclude methods that require substantially more training resources~\cite{zhang2025vsafastervideodiffusion} compared with \textsc{LinVideo} or support only specific video shapes~\cite{li2025radialattentiononlogn}. To be noted, in Sec.~\ref{sec:sla}, we compare our work with SLA~\cite{zhang2025slasparsitydiffusiontransformers}, which relies on a specific CUDA kernel implementation that only supports the RTX 5090 GPU.

\subsection{Main Results}\label{sec:results}

We compare our method with baselines and benchmark end-to-end latency across baselines using a batch size of $1$ with $50$ denoising steps in Tab.~\ref{tab:compare}. \textsc{LinVideo} consistently far outperforms all sparse-attention baselines and surpasses dense attention (FlashAttention2) on certain metrics, such as Overall Consistency and Scene Consistency. Moreover, our method achieves an average speedup of $1.43\text{-}1.71\times$, compared with $1.31\text{-}1.61\times$ for SVG~\cite{xi2025sparsevideogenacceleratingvideo}. It is also worth noting that for \textsc{LinVideo}, we do not employ any specialized kernel implementation, which, as a future direction, would allow our method to achieve an enhanced speedup ratio. Notably, when combined with DMD2~\cite{yin2024improved}, 1.3B \textsc{LinVideo} yields only a $\sim1\%$ performance drop while achieving a $15.92\times$ end-to-end speedup. Besides, our current design uses only dense linear attention or retains dense quadratic attention. This is orthogonal to sparse-attention methods. Thus, future work could integrate efficient sparse modules into our proposed \textsc{LinVideo} to further improve efficiency and performance.

Additionally, we also employ more challenging VBench-2.0~\cite{zheng2025vbench20advancingvideogeneration} to evaluate performance. We compare our \textsc{LinVideo} with the best-performing method SVG2 (see Tab.~\ref{tab:compare}) and the lossless baseline FlashAttention2. As shown in Fig.~\ref{fig:vbench2}, our method for \wan1.3B can achieve the same total score as FA2 and a much higher total score than SVG2. Moreover, our 4-step distilled 1.3B model incurs less than $3\%$ performance drops with higher scores on specific metrics like Human Identity and Multi-View Consistency compared with FA2. 

For qualitative results, we present a visualization in Fig.~\ref{fig:vis}, which demonstrates that our method achieves exceptionally high visual quality, even after 4-step distillation. More visual results can be found in the Appendix

\subsection{Ablation Studies}\label{sec:ablation}
We employ 1.3B \textsc{LinVideo} for ablation studies. $5$ dimensions of VBench are employed to evaluate performance. Unless specified, the settings are the same as those in Sec.~\ref{sec:experiments}.

\begin{table}[ht]\setlength{\tabcolsep}{1pt}
\vspace{-0.3in}
 \renewcommand{\arraystretch}{1.05}
  \centering
  \caption{Ablation results across different values of \texttt{target} (``\texttt{tar}''). We employ $\texttt{target}=16$ in 1.3B \textsc{LinVideo}.} 
  \vspace{-0.1in}
  \resizebox{\linewidth}{!}{
  \begin{tabu}[t!]{l|c|cccccccc}
\toprule
\multirow{2}{*}{\texttt{tar}} & \multirow{2}{*}{\makecell{Latency\\ (s)}$\downarrow$} &\multirow{2}{*}{\rotatebox{0}{\makecell{Imaging\\Quality}$\uparrow$}} &  \multirow{2}{*}{\rotatebox{0}{\makecell{Aesthetic\\Quality}$\uparrow$}} & \multirow{2}{*}{\rotatebox{0}{\makecell{Motion\\Smoothness}$\uparrow$}} & \multirow{2}{*}{\rotatebox{0}{\makecell{Dynamic \\Degree}$\uparrow$}} & \multirow{2}{*}{\rotatebox{0}{\makecell{Overall\\Consistency}$\uparrow$}}\\
 & & & & \\
 \midrule

 \midrule
 10 & 78.14&\underline{66.32} &  \underline{59.18} & \textbf{98.68} & \textbf{60.06}  & 26.35\\
12 &74.81 &\textbf{66.36} & 59.14 & \underline{98.57} & \underline{59.73}  & \textbf{26.65}\\
14 & 71.48 &66.17 & 58.88 & 98.34 & 59.67  & 26.29\\
\rowcolor{mycolor!30}16 & 68.26 &66.07 & \textbf{59.41} & 98.19 & 59.67  & \underline{26.52}\\
18 & 65.00 &65.84 & 58.32 & 97.78 & 58.63 &  26.08\\
20 & 61.68 &64.38 & 57.02 & 95.49 & 57.12 &  23.30\\

\bottomrule
\end{tabu}
}
    \label{tab:target}
    \vspace{-0.25in}
\end{table}
\noindent\textbf{Choice of \texttt{target}.} We investigate the effect of different \texttt{target} values on the video quality. As shown in Tab.~\ref{tab:target}, our results reveal that a larger \texttt{target} leads to greater acceleration but at the cost of performance degradation, and \textit{vice versa}. Specifically, we find that performance degrades slowly as \texttt{target} increases, remaining stable until $\texttt{tar}=18$, after which we observe a non-negligible drop.

\begin{table}[ht]\setlength{\tabcolsep}{1pt}
\vspace{-0.05in}
 \renewcommand{\arraystretch}{1.05}
  \centering
  \caption{Ablation results of \textit{selective transfer}. For \textsc{LinVideo}, $\lambda=0.01$. ``\textit{Manual}'' denotes we manually assign the same quadratic attention layers as \textsc{LinVideo} to linear attention, and ``\textit{Heuristic}'' signifies we employ a grid search method (details can be found in the Appendix) to determine the indices of quadratic attention layers to be replaced. Both methods employ $\mathcal{L}_\text{ADM}$ as their training loss after attention replacement.  We provide ablations for $\alpha$ of $\mathcal{L}_\text{reg}$ in the Appendix.} 
  \vspace{-0.1in}
  \resizebox{0.93\linewidth}{!}{
  \begin{tabu}[t!]{l|ccccc}
\toprule
\multirow{2}{*}{Method} & \multirow{2}{*}{\rotatebox{0}{\makecell{Imaging\\Quality}$\uparrow$}} & \multirow{2}{*}{\rotatebox{0}{\makecell{Aesthetic\\Quality}$\uparrow$}}  & \multirow{2}{*}{\rotatebox{0}{\makecell{Motion\\Smoothness}$\uparrow$}} & \multirow{2}{*}{\rotatebox{0}{\makecell{Dynamic \\Degree}$\uparrow$}} & \multirow{2}{*}{\rotatebox{0}{\makecell{Overall\\Consistency}$\uparrow$}}\\
 & &  & & \\
\midrule
  
\midrule
\rowcolor{mycolor!30}\textsc{LinVideo} & \underline{66.07} & \textbf{59.41} & \textbf{98.19} & \textbf{59.67} & \textbf{26.52} \\
\midrule
\textit{Manual} & 62.97 & 57.21 & 92.25 & 52.87 & 20.08 \\
\textit{Heuristic} & 60.74 & 54.13 & 90.36 & 50.61 & 18.94 \\
\midrule
$\lambda=0.1$ & \textbf{66.21} & \underline{59.17} & 97.94 & 59.31 & 26.16\\
$\lambda=0.001$ & 65.98 & 58.96 & \underline{98.14} & \underline{59.46} & \underline{26.37} \\
\midrule
w/o $\mathcal{L}_\text{reg}$ & 18.62 & 17.83 & 12.59 & 7.48 & 1.42 \\

\bottomrule
\end{tabu}
}
    \label{tab:select_trans}
    \vspace{-0.15in}
\end{table}
\noindent\textbf{Effect of selective transfer.} As shown in Tab.~\ref{tab:select_trans}, we study the effect of the proposed \textit{selective transfer}. \textsc{LinVideo} and \textit{Manual} replace the layers with indices $\{2\text{--}8,10\text{--}13,15\text{--}16,23,25,30\}$ with linear attention, while \textit{Heuristic} replaces $\{3\text{--}11,13\text{--}16,23,27,30\}$. In the table, \textit{Manual} clearly outperforms \textit{Heuristic}, indicating that our training-based layer selection is more effective than heuristic rules. \textsc{LinVideo} further improves upon \textit{Manual}, showing that the learnable score $r$ in Eq.~(\ref{eq:training_attention}) enables a progressive and stable conversion from quadratic to linear attention that benefits training. We also ablate the coefficient $\lambda$ of $\mathcal{L}_\text{con}+\mathcal{L}_\text{reg}$ in Eq.~(\ref{eq:first-loss}). Across metrics and $\lambda$ values, the performance variation is about $1\%$, indicating that \textsc{LinVideo} is not sensitive to $\lambda$. Finally, removing $\mathcal{L}_\text{reg}$ leads to a notable performance drop, confirming its role in improving training (\emph{i.e.}, reducing fluctuation of $r$) and eliminating rounding-induced errors (\emph{i.e.}, $\text{s.t.} |r-\lceil r\rfloor| <10^{-3}$ for each $r$). Visualization of its effect is also provided in Fig.~\ref{fig:rounding}.

\begin{table}[ht]\setlength{\tabcolsep}{1pt}
\vspace{-0.3in}
 \renewcommand{\arraystretch}{1.05}
  \centering
  \caption{Ablation results of ADM. ``w/ $\mathcal{L}_\text{mse}$'' and ``w/ $\mathcal{L}_\text{DMD}$'' denotes employ these two objectives to replace our $\mathcal{L}_\text{ADM}$ in Eq.~(\ref{eq:first-loss}), respectively. ``w/ $\hat s_t^\dagger$'' represents we employ $\mathcal{L}_\text{ADM}$ but train an additional model initialized from \wan 1.3B~\cite{wanteam2025wanopenadvancedlargescale}, which is to approximate the score function $\hat s_t$, at the same time. Following DMD~\cite{yin2024onestep}, we employ $5\times$ training iterations (\emph{i.e.}, $15K$) for the additional learned score approximator in both ``w/ $\mathcal{L}_\text{DMD}$'' and ``w/ $\hat s_t^\dagger$''.} 
  \vspace{-0.1in}
  \resizebox{0.93\linewidth}{!}{
  \begin{tabu}[t!]{l|ccccc}
\toprule
\multirow{2}{*}{Method} & \multirow{2}{*}{\rotatebox{0}{\makecell{Imaging\\Quality}$\uparrow$}} & \multirow{2}{*}{\rotatebox{0}{\makecell{Aesthetic\\Quality}$\uparrow$}}  &  \multirow{2}{*}{\rotatebox{0}{\makecell{Motion\\Smoothness}$\uparrow$}} & \multirow{2}{*}{\rotatebox{0}{\makecell{Dynamic \\Degree}$\uparrow$}} & \multirow{2}{*}{\rotatebox{0}{\makecell{Overall\\Consistency}$\uparrow$}} \\
 & &  & & \\
\midrule
  
\midrule
\rowcolor{mycolor!30}\textsc{LinVideo} & \textbf{66.07} & \textbf{59.41} & \textbf{98.19} & \textbf{59.67} & \textbf{26.52} \\
\midrule
w/ $\mathcal{L}_\text{mse}$ & 61.56 & 56.37 & 96.32 & 52.48 & 21.46 \\
w/ $\mathcal{L}_\text{DMD}$ & 57.44 & 52.79 & 90.72 & 49.37 & 16.96 \\
\midrule
w/ $\hat s_t^\dagger$ & \underline{65.61} & \underline{59.34} & \underline{97.82} & \underline{59.43} & \underline{25.87} \\
\bottomrule
\end{tabu}
}
    \label{tab:ADM}
\end{table}
\begin{figure}[ht!]
\vspace{-0.15in}
   \centering
    \setlength{\abovecaptionskip}{0.2cm}
        \includegraphics[width=0.45\textwidth]{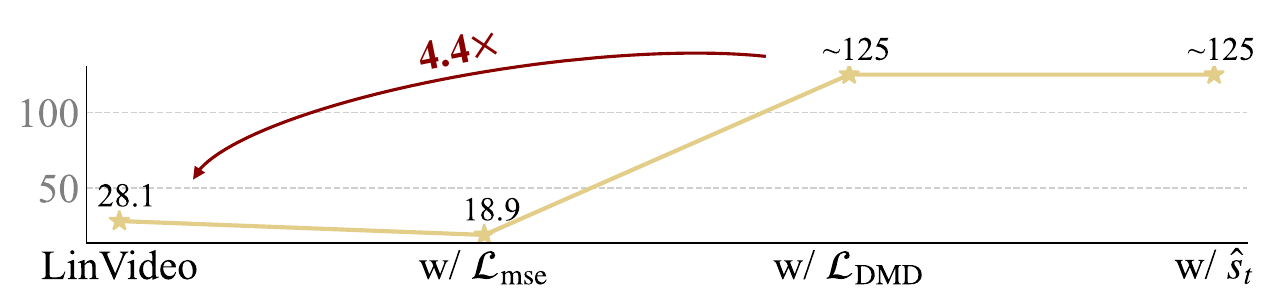}
        \vspace{-0.06in}
        \caption{Training hours across different objectives. Settings are the same as those in Tab.~\ref{tab:ADM}. FLOPs comparison can be found in Tab.~\ref{tab:flops}.}
        \vspace{-0.2in}
\end{figure}
\noindent\textbf{Effect of ADM.} We study several training objectives, as shown in Tab.~\ref{tab:ADM}. Our $\mathcal{L}_\text{ADM}$ outperforms the naive $\mathcal{L}_\text{mse}$ in Eq.~(\ref{eq:naive_objective}) and the few-step distribution matching loss $\mathcal{L}_\text{DMD}$ that aligns only $p_g$ and $p_0$ (see Sec.~\ref{sec:tdm}), with a clear margin. In addition, $\mathcal{L}_\text{ADM}$ reduces training time by $\sim4.4\times$ compared with $\mathcal{L}_\text{DMD}$ and the $\hat s_t^\dagger$ variant, both of which require training an extra model to estimate $\hat s_t$. Besides, our \textsc{LinVideo} transfers the model smoothly and progressively from quadratic- to linear-attention \textit{flow models}. Thus, it is reasonable to view $\hat\vu_\theta$ as a \textit{flow model} throughout this process. Experimentally, using $\hat\vu_\theta$ itself to compute $\hat s_t$ is more accurate, and thus achieves higher performance than $\hat s_t^\dagger$, which trains a separate score approximator.

\section{Conclusion}\label{sec:conclusions_and_limitations}
In this work, we explore accelerating the video diffusion model (DM) inference with linear attention in a post-training, \textit{data-free} manner. We first observe that replacing different quadratic attention layers with linear attention leads to large performance disparities. To address this, we propose \textit{selective transfer}, which automatically and progressively converts a target number of layers to linear attention while minimizing performance loss. Furthermore, finding that existing objectives are ineffective and inefficient in our scenarios, we introduce an \textit{anytime distribution matching} (ADM) objective that aligns sample distributions with the original model across all timesteps in the sampling trajectory. This substantially improves output quality. Extensive experiments demonstrate that \textsc{LinVideo} delivers lossless acceleration across different benchmarks.

\section*{Acknowledgement}
This work was supported by the National Natural Science Foundation of China (Nos. 62476018), and the Postdoctoral Fellowship Program of CPSF (No. BX20250487). This work was also supported by the Hong Kong Research Grants Council under the Areas of Excellence scheme grant AoE/E-601/22-R and NSFC/RGC Collaborative Research Scheme grant CRS\_HKUST603/22.

{
    \small
    \bibliographystyle{ieeenat_fullname}
    \bibliography{main}
}

\newpage
\appendix
\begin{center}{\bf \Large Appendix}\end{center}\vspace{-2mm}
\renewcommand{\thetable}{\Roman{table}}
\renewcommand{\thefigure}{\Roman{figure}}
\renewcommand{\theequation}{\Roman{equation}}
\setcounter{table}{0}
\setcounter{figure}{0}
\setcounter{equation}{0}

\Crefname{appendix}{Appendix}{Appendixes}

This document supplements the main paper as follows:
\begin{itemize}[leftmargin=*]
    \item Sec.~\ref{sec:more-imple} provides more implementation details;
    \item Sec.~\ref{sec:alpha} details the effect of hyperparameter $\alpha$ of Eq.~(\ref{eq:reg});
    \item Sec.~\ref{sec:lreg} provides additional ablation studies for $\alpha$;
    \item Sec.~\ref{sec:derivation-score} gives the derivation of score difference (\emph{i.e.}, Eq.~(\ref{eq:score}));
    \item Sec.~\ref{sec:temporal-jitter} presents examples of temporal jitter incurred by naive objective (\emph{i.e.}, Eq.~(\ref{eq:naive_objective}));
     \item Sec.~\ref{sec:combine} shows the results of directly distilling the original DM into a 4-step model with linear attention;
    \item Sec.~\ref{sec:details-heuristic} provides detailed setups for heuristic search in Tab.~\ref{tab:select_trans};
    \item Sec.~\ref{sec:comp-few} provides additional comparison for DMD2 distilled few-step models;
    \item Sec.~\ref{sec:kernel} presents ablation studies for the kernel function of linear attention;
    \item Sec.~\ref{sec:sla} provides a comparison between SLA and our method;
    \item Sec.~\ref{sec:cog} shows the comparison results for CogVideoX~\cite{yang2025cogvideoxtexttovideodiffusionmodels};
    \item Sec.~\ref{sec:vis} presents visualization examples of our \textsc{LinVideo}.
\end{itemize}

\section{More Implementation Details}\label{sec:more-imple}
In this section, we provide additional implementation details. Before training, we first collect inputs and outputs from \wan~\cite{wanteam2025wanopenadvancedlargescale} across different sampling steps using prompts from \texttt{OpenVid}~\cite{nan2024openvid1m}. Each input/output contains $81$ frames ($21$ frames in latent space), with an original resolution of $480p$ for \wan 1.3B and $720p$ for \wan 14B. During training, we use \texttt{Pytorch FSDP}~\cite{zhao2023pytorchfsdpexperiencesscaling} with a warm-up phase covering $\frac{1}{10}$ of the total training epochs, and set the global batch size to $48$ for \wan 1.3B and $64$ for \wan 14B. The learning rate follows a cosine annealing schedule, starting from $1\times10^{-4}$. For other hyperparameters, we set $\lambda = 0.01$, and $\alpha$ follows a linear decay from $20$ to $2$. For evaluation, we sample $5$ videos for each unaugmented text prompt across the $8$ dimensions in VBench~\cite{huang2024vbench}. For VBench-2.0~\citep{zheng2025vbench20advancingvideogeneration}, we generate $3$ videos per augmented text prompt across all dimensions, except for the Diversity dimension, where we generate $20$ videos for each prompt.

\section{Effect of \texorpdfstring{$\alpha$}{alpha}}\label{sec:alpha}
\begin{figure}[ht!]
\vspace{-0.1in}
   \centering
    \setlength{\abovecaptionskip}{0.2cm}
        \includegraphics[width=0.40\textwidth]{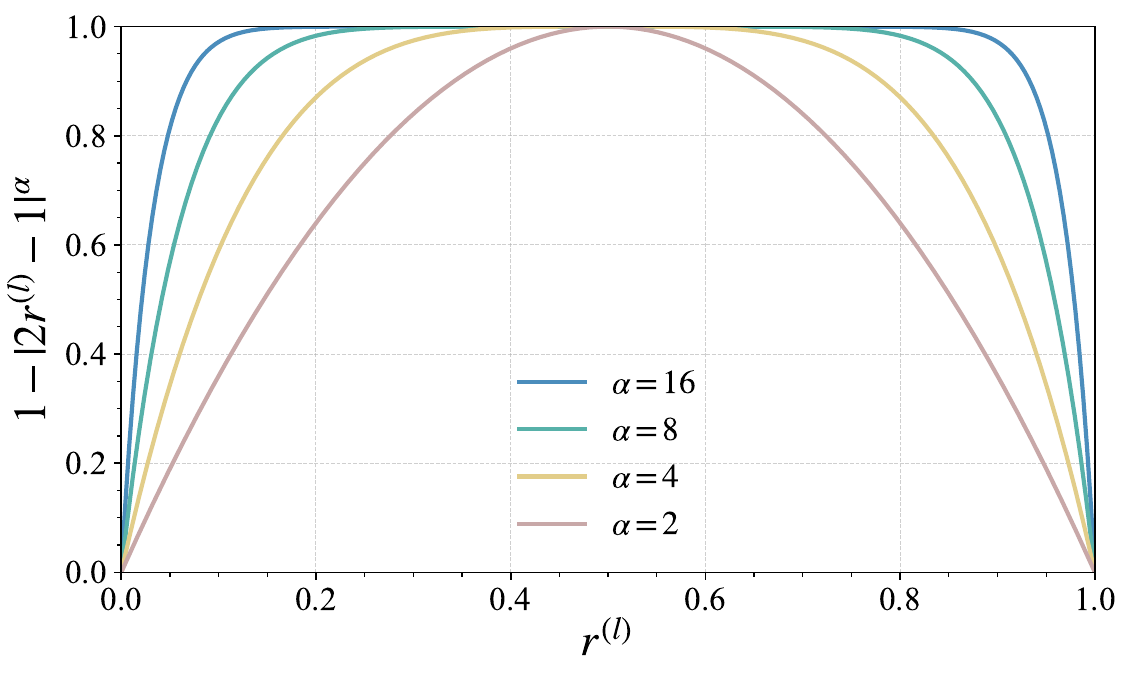}
        \vspace{-0.1in}
        \caption{Effect of $\alpha$ on $1-|2r^{(l)}-1|^\alpha$ of Eq.~(\ref{eq:reg}).}
        \label{fig:alpha}
        \vspace{-0.1in}
\end{figure}

As mentioned in the main text, we annealing decay the parameter $\alpha$ from large to small. This encourages $r$ to move more adaptively at the initial phase to improve the training loss, but forces it to $0.0/1.0$ in the later phase. To understand this behavior, we provide a visualization in Fig.~\ref{fig:alpha}. As shown in this figure, when $\alpha$ is large, the function remains nearly constant (close to $1.0$) from $r^{(l)} = 0.5$ extending towards both boundaries, resulting in near-zero gradients in most regions. This flat landscape allows $r$ to explore more freely during initial training. Conversely, when $\alpha$ is small, the gradient remains significant across a broader range, effectively pushing $r$ towards the boundaries ($0.0$ or $1.0$) in later training stages.

\section{Ablation for \texorpdfstring{$\alpha$}{alpha}}\label{sec:lreg}
Here, we provide additional ablation studies for $\alpha$ of $\mathcal{L}_\text{reg}$. As demonstrated in Tab.~\ref{tab:alpha}, when we narrow or expand the range of $\alpha$, the variation in all metrics stays within $0.03$, which shows that our loss is not sensitive to changes in the chosen $\alpha$ values. In addition, this range of $\alpha$ yields strong final performance for models from 1.3B to 14B (Tab.~\ref{tab:compare}), demonstrating good generalization. We also verify experimentally that using a dynamically changing $\alpha$ gives non-negligible improvements compared with using a fixed $\alpha$, which confirms the benefit of our design: $\alpha$ encourages free exploration in the early stage of training and is gradually guided toward $0.0/1.0$ only in the later stage.

\begin{table}[ht]\setlength{\tabcolsep}{1pt}
\vspace{-0.05in}
 \renewcommand{\arraystretch}{1.05}
  \centering
  \caption{Ablation results of $\alpha$. Our \textsc{LinVideo} employs $20\rightarrow2$ as the range of $\alpha$.} 
  \vspace{-0.1in}
  \resizebox{0.93\linewidth}{!}{
  \begin{tabu}[t!]{l|ccccc}
\toprule
\multirow{2}{*}{Range of $\alpha$} & \multirow{2}{*}{\rotatebox{0}{\makecell{Imaging\\Quality}$\uparrow$}} & \multirow{2}{*}{\rotatebox{0}{\makecell{Aesthetic\\Quality}$\uparrow$}}  &  \multirow{2}{*}{\rotatebox{0}{\makecell{Motion\\Smoothness}$\uparrow$}} & \multirow{2}{*}{\rotatebox{0}{\makecell{Dynamic \\Degree}$\uparrow$}} & \multirow{2}{*}{\rotatebox{0}{\makecell{Overall\\Consistency}$\uparrow$}} \\
 & &  & & \\
\midrule
\midrule
\rowcolor{mycolor!30}$20\rightarrow 2$ & \underline{66.07} & \textbf{59.41} & \underline{98.19} & \underline{59.67} & \textbf{26.57}\\
\midrule
$16\rightarrow 4$ & \textbf{66.09} & 59.38 & 98.16 & \textbf{59.70} & 26.54\\
$24\rightarrow 1$ & 66.06 & \underline{59.40} & \textbf{98.20} & 59.65 & \underline{26.56}\\
\midrule
$\,\,\,4\rightarrow4$ & 65.87 & 59.08 & 98.10 & 59.08 & 26.27 \\

\bottomrule
\end{tabu}
}
    \label{tab:alpha}
\end{table}

\section{Derivation of Eq.~(\ref{eq:score})}\label{sec:derivation-score}
In this section, we provide a derivation of Eq.~(\ref{eq:score}). Following Ma \emph{et al.}~\cite{ma2024sitexploringflowdiffusionbased} (their Eq.~(9)), the score $s_t$ associated with $\hat{\bx}_t$ and estimated by $\vu_\theta$ can be written as
\begin{equation}
s_t(\hat{\bx}_t)
= \sigma_t^{-1}\,\frac{\alpha_t \vu_\theta(\hat{\bx}_t) - \dot{\alpha}_t \hat{\bx}_t}{\dot{\alpha}_t \sigma_t - \alpha_t \dot{\sigma}_t},
\end{equation}
where $\alpha_t, \sigma_t$ are the noise schedules and $\dot{\alpha}_t, \dot{\sigma}_t$ are their first-order derivatives with respect to $t$. For the \textit{rectified flow models}~\cite{wanteam2025wanopenadvancedlargescale} considered in this work, we take $\alpha_t = 1 - t$ and $\sigma_t = t$. Substituting these schedules into the above expression yields
\begin{equation}
s_t(\hat{\bx}_t)
= \frac{1}{t}\frac{(1-t)\vu_\theta(\hat{\bx}_t)+\hat{\bx}_t}{-t-(1-t)}
= -\frac{1}{t}\big((1-t)\vu_\theta(\hat{\bx}_t)+\hat{\bx}_t\big).
\end{equation}
Likewise, the score $\hat{s}_t$ estimated by $\hat{\vu}_\theta$ corresponding to $\hat{\bx}_t$ is given by
\begin{equation}
\hat{s}_t(\hat{\bx}_t)
= -\frac{1}{t}\big((1-t)\hat{\vu}_\theta(\hat{\bx}_t)+\hat{\bx}_t\big).
\end{equation}
Therefore, the score difference is
\begin{equation}
s_t(\hat{\bx}_t) - \hat{s}_t(\hat{\bx}_t)
= -\frac{1-t}{t}\,\big(\vu_\theta(\hat{\bx}_t) - \hat{\vu}_\theta(\hat{\bx}_t)\big).
\end{equation}

\section{Temporal Jitter of Naive Objective}\label{sec:temporal-jitter}
\begin{figure}[ht!]
    \centering
    \begin{minipage}[b]{\linewidth}
       \begin{minipage}[b]{\linewidth}
            \centering
            \begin{subfigure}[tp!]{\textwidth}
            \centering
            \includegraphics[width=\linewidth]{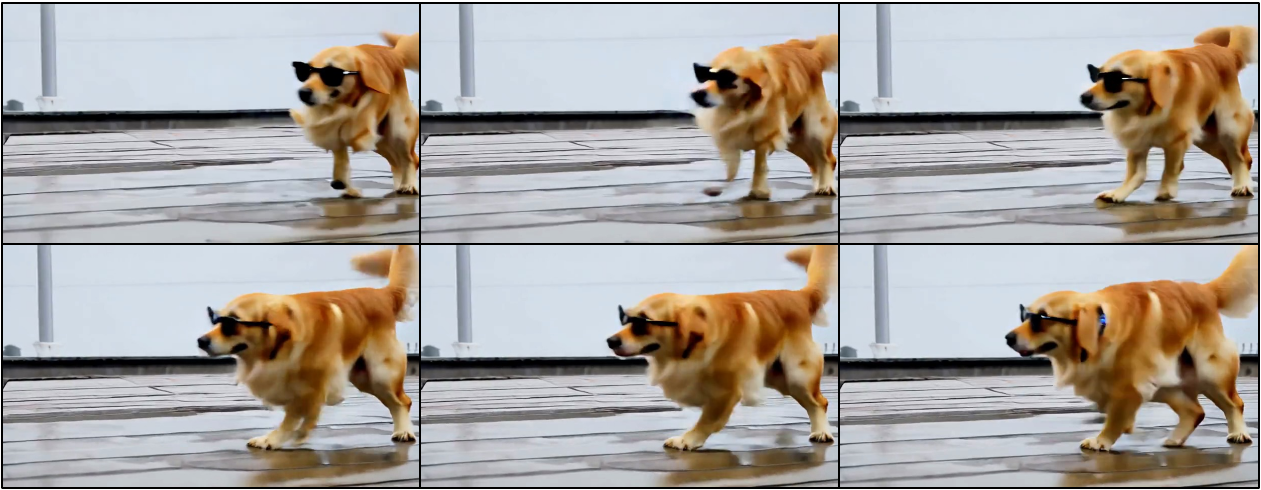}
            \end{subfigure}
       \end{minipage}\hfill
       \begin{minipage}[b]{\linewidth}
            \centering
            \begin{subfigure}[tp!]{\linewidth}
            \centering
            \includegraphics[width=\linewidth]{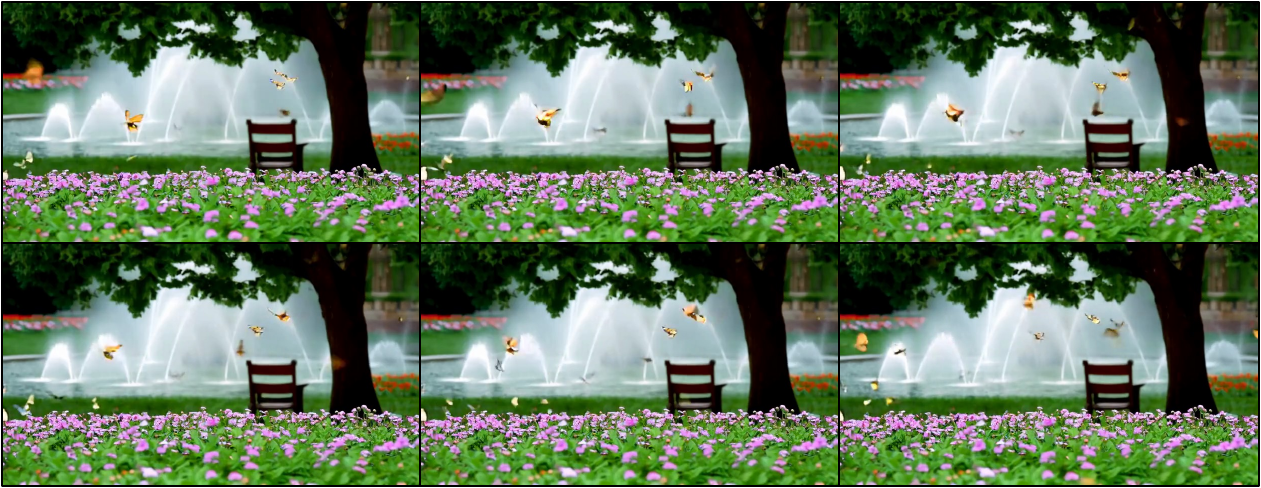}
            \end{subfigure}
       \end{minipage}
   \end{minipage}
    \vspace{-0.1in}
    \caption{A segment of $6$ adjacent frames from an $81$-frame, $480p$, $16$ fps video generated by the linear-attention \wan 1.3B~\cite{wanteam2025wanopenadvancedlargescale} trained with the naive objective. Other settings are the same as those for \textsc{LinVideo}. Both the dog’s legs (\textit{Upper}) and the butterflies (\textit{Lower}) exhibit severe temporal jitter.}
    \label{fig:temporl}
    \vspace{-0.1in}
\end{figure}

We provide examples in Fig.~\ref{fig:temporl} using the naive training objective $\mathcal{L}_\text{mse}$ (\emph{i.e.}, Eq.~(\ref{eq:naive_objective})), where the results clearly exhibit temporal jitter.

\section{Directly combine Few-step Distillation with Linear Attention}\label{sec:combine}
As shown in Tab.~\ref{tab:direct}, we observe that directly combining a few-step distillation with linear attention replacement (\emph{i.e.}, our \textit{selective transfer}) leads to severe performance degradation. This motivates a two-stage pipeline, \emph{viz.}, \textsc{LinVideo}$\rightarrow$DMD2~\cite{yin2024improved}. In the first stage, linear attention is introduced and already attains strong performance; in the second stage, few-step distillation is applied to further accelerate the model. This two-stage design stabilizes training and avoids the collapse observed with the direct combination.
\begin{table}[ht]\setlength{\tabcolsep}{1pt}
\vspace{-0.05in}
 \renewcommand{\arraystretch}{1.05}
  \centering
  \caption{Performance for the few-step distilled linear attention \wan 1.3B~\cite{wanteam2025wanopenadvancedlargescale}. ``\textsc{LinVideo} + DMD2'' denotes that we first employ \textsc{LinVideo} to obatin a linear attention DM and then use DMD2~\cite{yin2024improved} to distill is to the 4-step version. ``\textit{ST} + DMD2'' implies that we combine our \textit{selective transfer} and DMD2 to obtain the 4-step linear attention model in a single training stage.} 
  \vspace{-0.1in}
  \resizebox{\linewidth}{!}{
  \begin{tabu}[t!]{l|ccccc}
\toprule
\multirow{2}{*}{Method} & \multirow{2}{*}{\rotatebox{0}{\makecell{Imaging\\Quality}$\uparrow$}} & \multirow{2}{*}{\rotatebox{0}{\makecell{Aesthetic\\Quality}$\uparrow$}}  &  \multirow{2}{*}{\rotatebox{0}{\makecell{Motion\\Smoothness}$\uparrow$}} & \multirow{2}{*}{\rotatebox{0}{\makecell{Dynamic \\Degree}$\uparrow$}} & \multirow{2}{*}{\rotatebox{0}{\makecell{Overall\\Consistency}$\uparrow$}} \\
 & &  & & \\
\midrule
\midrule
\rowcolor{mycolor!30}\textsc{LinVideo} + DMD2 & \textbf{65.62} & \textbf{57.74} & \textbf{97.32} & \textbf{61.26} & \textbf{25.94}\\
\textit{ST} + DMD2 & 48.72 & 39.64 & 73.85 & 41.97 & 11.06\\

\bottomrule
\end{tabu}
}
    \label{tab:direct}
\end{table}

\section{Details of Heuristic Search}\label{sec:details-heuristic}
For the \textit{Heuristic} search method in the ablation study, we use $128$ prompts from \texttt{OpenVid}~\cite{nan2024openvid1m}. Starting from the original DM, at each step we consider all remaining quadratic attention layers, construct modified variants where only one candidate layer is replaced by a linear attention layer, and measure the resulting output differences. We then permanently replace the layer that yields the smallest difference and repeat this procedure based on the updated model until the desired number of \texttt{target} layers has been selected.

\section{Comparison for Few-step Models}\label{sec:comp-few}
Tab.~\ref{tab:dmd}~\footnote{We report the average score of the chosen VBench dimensions Tab.~\ref{tab:compare}. ``FA2'' denotes lossless FlashAttention2 baseline.} shows that, when combined with DMD2, our method outperforms all baselines.

\begin{table}[!t]\setlength{\tabcolsep}{1pt}
  \centering
  \begin{minipage}[t]{0.69\linewidth}
  \vspace{0pt}\centering
  \renewcommand{\arraystretch}{1.}
    \caption{Comparison on \texttt{Wan} 1.3B (all methods use 4-step DMD2). Due to resource limits, we will add more results in the future.}
    \vspace{-0.2in}
    \vspace{0.04in}
      \resizebox{\linewidth}{!}{
      \begin{tabu}[t!]{l|c|ccccc}
\toprule
Metric & FA2 & DFA & XAttn & SVG & SVG2 & \cellcolor{mycolor!30}\textsc{LinVideo}\\
\midrule 
VBench$\uparrow$ & 67.68 & 64.71 & 64.32 & 65.83 & \underline{66.06} & \cellcolor{mycolor!30}\textbf{66.98}\\
Latency (s)$\downarrow$ & 8.76 & 8.26 & 7.82 & \underline{6.95} & 7.96 & \cellcolor{mycolor!30}\textbf{6.11}\\
\bottomrule
\end{tabu}
}
        \label{tab:dmd}
  \end{minipage}\hfill
  \begin{minipage}[t]{0.3\linewidth}
  \vspace{0pt}\centering
  \renewcommand{\arraystretch}{1.}
    \caption{Training FLOPs ($3K$ steps).}
    \vspace{-0.14in}
      \resizebox{0.9\linewidth}{!}{
      \begin{tabu}[t!]{l|c}
\toprule
Method & FLOPs$\downarrow$ \\
\midrule
\rowcolor{mycolor!30}\textsc{LinVideo} & \underline{$5.97{\times}10^4$P}\\
w/ $\mathcal{L}_\text{mse}$ & \textbf{$\mathbf{4.77{\times}10^4}$P}\\
w/ $\mathcal{L}_\text{DMD}$ & $2.97{\times}10^5$P\\
w/ $\hat{s}_t$ & $2.97{\times}10^5$P \\
\bottomrule
\end{tabu}
}
        \label{tab:flops}
  \end{minipage}\hfill
\end{table}

\begin{table}[!ht]\setlength{\tabcolsep}{1pt}
  \centering

  \begin{minipage}[b]{\linewidth}
    \renewcommand{\arraystretch}{1.}

    \begin{minipage}[t]{0.31\linewidth}
    \vspace{0pt}
    \vspace{0.05in}
    \captionsetup{justification=raggedright,singlelinecheck=false}
    \captionof{table}{Comparison on CogVideoX-2B.}
    \label{tab:cogvideox}
  \end{minipage}\hfill
  \begin{minipage}[t]{0.69\linewidth}
    \vspace{0pt}
    \centering
    \vspace{0.04in}
    \resizebox{\linewidth}{!}{\begin{tabu}[t!]{l|c|ccccc}
\toprule
Metric & FA2 & DFA & XAttn & SVG & SVG2 & \cellcolor{mycolor!30}\textsc{LinVideo}\\
\midrule 
VBench$\uparrow$ & 65.97 & 65.36 & 65.08 & 65.54 & \underline{65.78} & \cellcolor{mycolor!30}\textbf{65.97}\\
Latency (s)$\downarrow$ & 41.35 & 37.84 & 34.29 & \underline{33.13} & 38.46 & \cellcolor{mycolor!30}\textbf{29.64}\\
\bottomrule
\end{tabu}
}
  \end{minipage}
  \vspace{-0.05in}
  \end{minipage}
  \begin{minipage}[b]{0.58\linewidth}
    \renewcommand{\arraystretch}{1.}
    \caption{Comparison with SLA. ``*'' denotes results on an RTX5090 GPU; other results are on an H100 GPU.}
    \vspace{-0.2in}
    \vspace{0.04in}
    \resizebox{\linewidth}{!}{\begin{tabu}[t!]{l|ccc}
\toprule
Method & VBench$\uparrow$ & Latency (s)$\downarrow$ & Latency (s)$\downarrow^*$\\
\midrule
\texttt{Wan} 1.3B & 67.63 & 97.32 & 162.65\\
\midrule
SLA & \underline{65.72} & \underline{93.95} & \textbf{78.42}\\
\rowcolor{mycolor!30}\textsc{LinVideo} & \textbf{67.61} & \textbf{68.26} & \underline{112.14}\\
\bottomrule
\end{tabu}
}
    \label{tab:sla}
  \end{minipage}
  \begin{minipage}[b]{0.41\linewidth}
    \renewcommand{\arraystretch}{1.1}
    \caption{Comparison across kernel functions on \texttt{Wan} 1.3B.}
    \vspace{-0.16in}
    \resizebox{\linewidth}{!}{\begin{tabu}[t!]{l|cc}
\toprule
Kernel & VBench$\uparrow$ & Latency (s)$\downarrow$ \\
\midrule
\rowcolor{mycolor!30}\textit{Hedgehog} & \textbf{67.61} & \underline{68.26} \\
\texttt{ReLU} & 65.48 & \textbf{68.12} \\
Taylor Exp & \underline{67.24} & 68.54\\
\bottomrule
\end{tabu}
}
    \label{tab:kernel}
  \end{minipage}\hfill

\end{table}

\section{Ablation for Kernel Function}\label{sec:kernel}
\textit{Hedgehog}~\cite{zhang2024the} preserves softmax-like \textit{spiky weights} and \textit{dot-product monotonicity}, making it more expressive than typical linear attention kernels. This benefits video generation (verify in Tab.~\ref{tab:kernel}~\footnote{Taylor Exp: refer to Zhang \emph{et al.}~\cite{zhang2024the}.}), where long sequences require sharp, stable long-range interactions.

\section{Comparison between \textsc{LinVideo} and SLA}\label{sec:sla}
As shown in Tab.~\ref{tab:sla}, \textsc{LinVideo} outperforms SLA on VBench. SLA’s large speedup mainly comes from its hardware-specific kernels (currently only supporting RTX5090), while \textsc{LinVideo} uses \texttt{torch} implementation. Moreover, \textsc{LinVideo} can be further accelerated if we combine sparse attention as SLA ($95\%$ sparsity).

\section{Comparison for CogVideoX}\label{sec:cog}
Our methods do not rely on any specific model architecture design. As shown in Tab.~\ref{tab:cogvideox}, \textsc{LinVideo} outperforms baselines on CogVideoX.

\section{Visualization Results}\label{sec:vis}
\begin{figure*}[!ht]
\vspace{-0.1in}
   \centering
   \setlength{\abovecaptionskip}{0.2cm}
    \begin{minipage}[b]{\linewidth}
        \begin{spacing}{1}
       {\footnotesize Prompt: \textit{``A volcano erupts in the distance, glowing lava rivers flowing against a darkened sky.''}}
    \end{spacing}
       \begin{minipage}[b]{\linewidth}
            \centering
            \begin{subfigure}[tp!]{\textwidth}
            \centering
            \includegraphics[width=\linewidth]{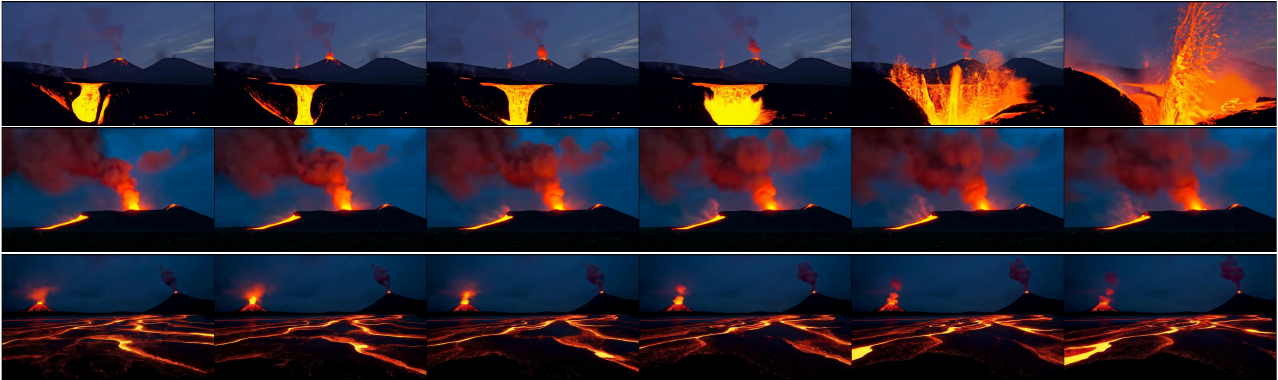}
            \end{subfigure}
       \end{minipage}
       \vspace{0.02in}
       \begin{spacing}{1}
       {\footnotesize Prompt: \textit{``A pod of dolphins leaps out of the sparkling ocean in graceful arcs, splashing back into the water as the horizon glows with sunset; the camera follows from the side, keeping a continuous rhythm with their motion.''}}
    \end{spacing}
       \begin{minipage}[b]{\linewidth}
            \centering
            \begin{subfigure}[tp!]{\textwidth}
            \centering
            \includegraphics[width=\linewidth]{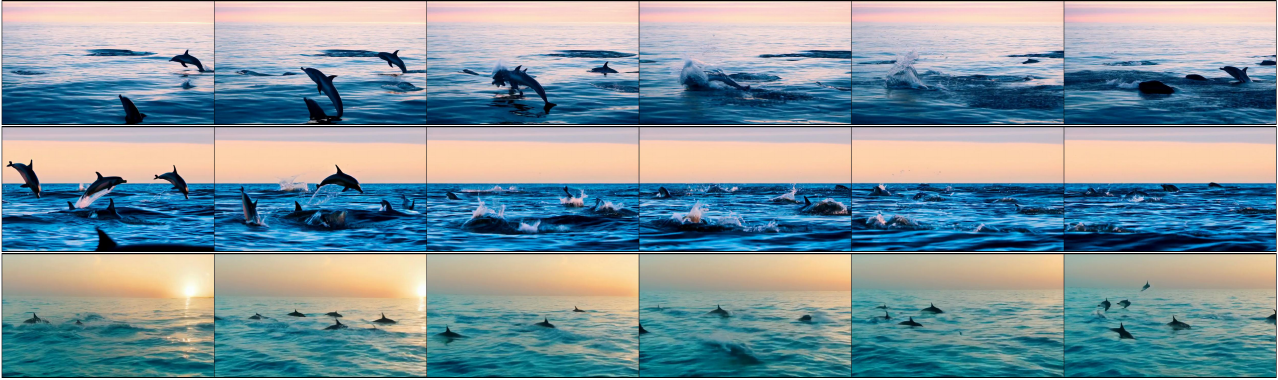}
            \end{subfigure}
       \end{minipage}
       \vspace{0.02in}
       \begin{spacing}{1}
       {\footnotesize Prompt: \textit{``A massive elephant walks slowly across a sunlit savannah, dust rising around its feet, the warm glow of sunset illuminating the horizon; the camera moves steadily forward alongside, emphasizing the grandeur of its stride.''}}
    \end{spacing}
       \begin{minipage}[b]{\linewidth}
            \centering
            \begin{subfigure}[tp!]{\linewidth}
            \centering
            \includegraphics[width=\linewidth]{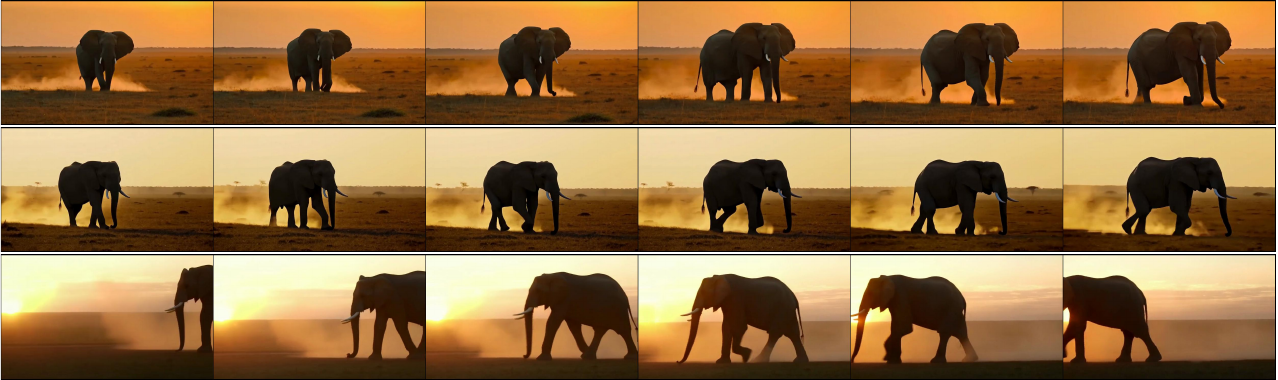}
            \end{subfigure}
       \end{minipage}
   \end{minipage}
   \caption{Visual results at $480p$ across \wan 1.3B~\cite{wanteam2025wanopenadvancedlargescale} (\textit{Upper}), 1.3B \textsc{LinVideo} (\textit{Middle}), and 1.3B \textsc{LinVideo}+ 4-step DMD2~\cite{yin2024improved} (\textit{Lower}).}
   \label{fig:1_3}
\end{figure*}

\begin{figure*}[!ht]
\vspace{-0.1in}
   \centering
   \setlength{\abovecaptionskip}{0.2cm}
    \begin{minipage}[b]{\linewidth}
        \begin{spacing}{1}
       {\footnotesize Prompt: \textit{``Retro 80s Monster Horror Comedy Movie Scene: Color film, children's bedroom bathed in soft, warm light. Plush monsters of various sizes and colors are having a chaotic party, jumping on the bed, dancing to upbeat music, and throwing confetti. The walls are adorned with posters of classic 80s movies, and the room is filled with the playful laughter of children.''}}
    \end{spacing}
       \begin{minipage}[b]{\linewidth}
            \centering
            \begin{subfigure}[tp!]{\textwidth}
            \centering
            \includegraphics[width=\linewidth]{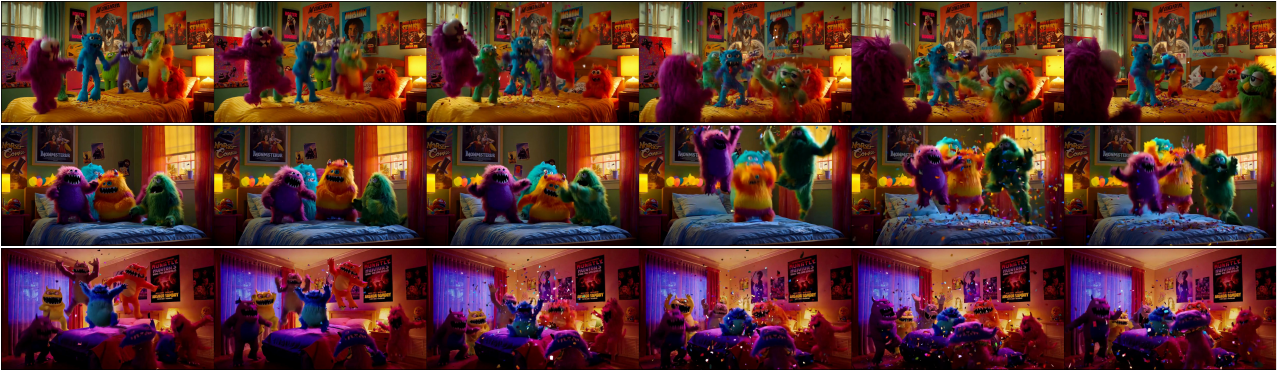}
            \end{subfigure}
       \end{minipage}
       \vspace{0.02in}
       \begin{spacing}{1}
       {\footnotesize Prompt: \textit{``A drone camera circles around a beautiful historic church built on a rocky outcropping along the Amalfi Coast, the view showcases historic and magnificent architectural details and tiered pathways and patios, waves are seen crashing against the rocks below as the view overlooks the horizon of the coastal waters and hilly landscapes of the Amalfi Coast Italy, several distant people are seen walking and enjoying vistas on patios of the dramatic ocean views, the warm glow of the afternoon sun creates a magical and romantic feeling to the scene, the view is stunning captured with beautiful photography.''}}
    \end{spacing}
       \begin{minipage}[b]{\linewidth}
            \centering
            \begin{subfigure}[tp!]{\textwidth}
            \centering
            \includegraphics[width=\linewidth]{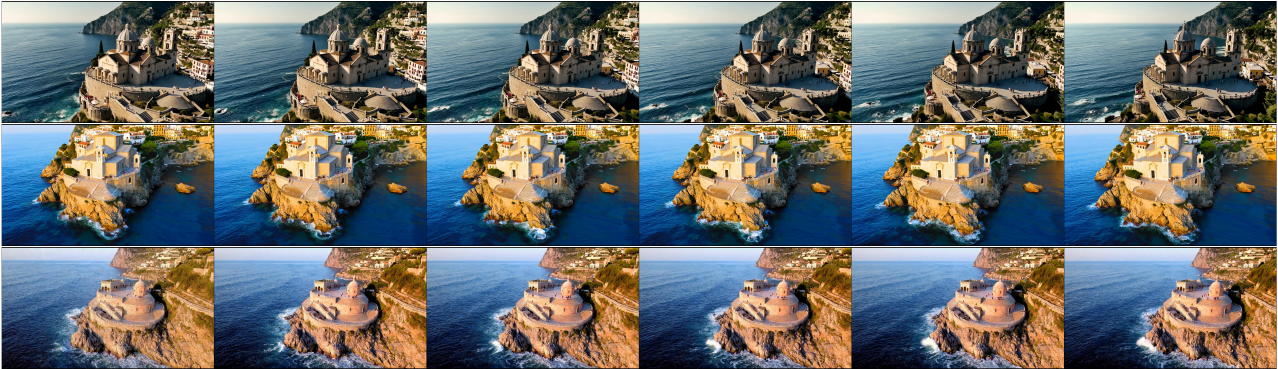}
            \end{subfigure}
       \end{minipage}
       \vspace{0.02in}
       \begin{spacing}{1}
       {\footnotesize Prompt: \textit{``An extreme close-up of an gray-haired man with a beard in his 60s, he is deep in thought pondering the history of the universe as he sits at a cafe in Paris, his eyes focus on people offscreen as they walk as he sits mostly motionless, he is dressed in a wool coat suit coat with a button-down shirt, he wears a brown beret and glasses and has a very professorial appearance, and the end he offers a subtle closed-mouth smile as if he found the answer to the mystery of life, the lighting is very cinematic with the golden light and the Parisian streets and city in the background, depth of field, cinematic 35mm film.''}}
    \end{spacing}
       \begin{minipage}[b]{\linewidth}
            \centering
            \begin{subfigure}[tp!]{\linewidth}
            \centering
            \includegraphics[width=\linewidth]{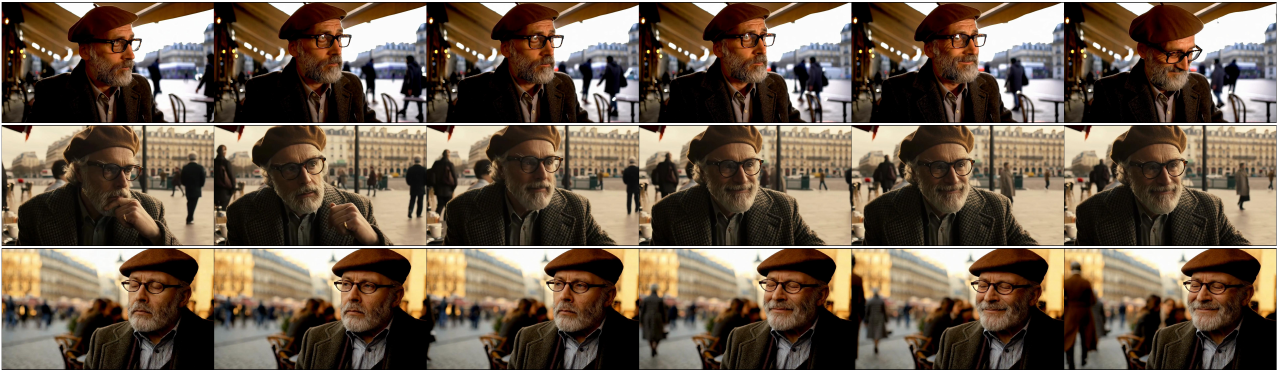}
            \end{subfigure}
       \end{minipage}
   \end{minipage}
   \caption{Visual results at $720p$ across \wan 14B~\cite{wanteam2025wanopenadvancedlargescale} (\textit{Upper}), 14B \textsc{LinVideo} (\textit{Middle}), and 14B \textsc{LinVideo}+ 4-step DMD2~\cite{yin2024improved} (\textit{Lower}).}
   \label{fig:14}
\end{figure*}

In this section, we present randomly selected samples generated by \textsc{LinVideo} without any cherry-picking, as shown in Figs.~\ref{fig:1_3}--\ref{fig:14}. For a more detailed inspection, we recommend zooming in to closely examine the individual frames.

\end{document}